\documentclass[10pt,journal,compsoc]{IEEEtran}
\usepackage[utf8]{inputenc}
\usepackage{epsfig}
\usepackage{graphicx}

\usepackage{amsmath}
\usepackage{amssymb}
\usepackage{tabularx}
\usepackage{lscape}
\usepackage{longtable}
\usepackage{wrapfig}

\usepackage{dirtytalk}

\DeclareRobustCommand{\hlorange}[1]{{\sethlcolor{orange}\hl{#1}}}

\usepackage{biblatex} 
\usepackage{rotating}
\usepackage{afterpage}
\usepackage{comment}
\usepackage{color,soul}
\usepackage{tikz}
\usetikzlibrary{positioning}
\usepackage{tablefootnote}

\newcommand{\subf}[2]{%
  {\small\begin{tabular}[t]{@{}c@{}}
   \mbox{}\\[-\ht\strutbox]
   #1\\#2
   \end{tabular}}%
}

\usepackage{pgfplots}
\pgfplotsset{width=7cm, compat=1.9}
\usepackage{pgf-pie}

\newlength\Textht
\usetikzlibrary{trees}

\usepackage{subcaption}
\usepackage{xcolor}
\usepackage{threeparttable}
\usepackage{algorithm}
\usepackage{algorithmic}
\usepackage{enumitem}
\usepackage{multirow}
\usepackage{lipsum}
\usepackage[hidelinks,pagebackref=false,breaklinks=true,colorlinks,bookmarks=false]{hyperref}
\ifCLASSINFOpdf
\else
\fi

\usepackage{comment}

\usepackage{xspace}
\makeatletter
\DeclareRobustCommand\onedot{\futurelet\@let@token\@onedot}
\def\@onedot{\ifx\@let@token.\else.\null\fi\xspace}

\newcolumntype{P}[1]{>{\centering\arraybackslash}p{#1}}

\makeatletter
\renewcommand*{\p@section}{\S\,}
\renewcommand*{\p@subsection}{\S\,}
\makeatother

\newcommand{\erhao}[1]{\fontsize{11pt}{\baselineskip}\selectfont}

\usepackage{url}
\usepackage[utf8]{inputenc}
\usepackage{booktabs}
\usepackage[american]{babel}
\usepackage[T1]{fontenc}
\usepackage{balance}

\begin{document}
\title{A Survey on Physics Informed Reinforcement Learning: Review and Open Problems }

\author{C.~Banerjee,~\IEEEmembership{Member,~IEEE,}
        K.~Nguyen,~\IEEEmembership{ Member,~IEEE,}
        C.~Fookes,~\IEEEmembership{Senior Member,~IEEE,}  
        M.~Raissi,~\IEEEmembership{Senior Member,~IEEE,}  
\IEEEcompsocitemizethanks{\IEEEcompsocthanksitem C. Banerjee, K. Nguyen, and C. Fookes are with Queensland University of Technology, Australia. Maziar Raissi is with University of Colorado Boulder, USA. \protect 
E-mail: \{c.banerjee, k.nguyenthanh, c.fookes\}@qut.edu.au, maziar.raissi@colorado.edu}


}

%
%

\markboth{}%
{Author \MakeLowercase{\textit{et al.}}: Towards Physics-Informed Computer Vision: A Review}

\IEEEtitleabstractindextext{%
\begin{abstract}
The inclusion of physical information in machine learning frameworks has revolutionized many application areas. This involves enhancing the learning process by incorporating physical constraints and adhering to physical laws. In this work we explore their utility for reinforcement learning applications. We present a thorough review of the literature on incorporating physics information, as known as physics priors, in reinforcement learning approaches, commonly referred to as physics-informed reinforcement learning (PIRL).
We introduce a novel taxonomy with the reinforcement learning pipeline as the backbone to classify existing works, compare and contrast them, and derive crucial insights.
Existing works are analyzed with regard to the representation/ form of the governing physics modeled for integration, their specific contribution to the typical reinforcement learning architecture, and their connection to the underlying reinforcement learning pipeline stages. We also identify core learning architectures and physics incorporation biases (i.e. observational, inductive and learning) of existing PIRL approaches and use them to further categorize the works for better understanding and adaptation. By providing a comprehensive perspective on the implementation of the physics-informed capability, the taxonomy presents a cohesive approach to PIRL. It identifies the areas where this approach has been applied, as well as the gaps and opportunities that exist. Additionally, the taxonomy sheds light on unresolved issues and challenges, which can guide future research. This nascent field holds great potential for enhancing reinforcement learning algorithms by increasing their physical plausibility, precision, data efficiency, and applicability in real-world scenarios.
\end{abstract}

\begin{IEEEkeywords}
Physics-informed, Reinforcement Learning, Machine learning, Neural Network, Deep Learning 
\end{IEEEkeywords}}

\maketitle

\IEEEdisplaynontitleabstractindextext

\IEEEpeerreviewmaketitle

\section{Introduction}



Through trial-and-error interactions with the environment,  Reinforcement Learning (RL) offers a promising approach to solving decision-making and optimization problems. Over the past few years, RL has accomplished impressive feats in handling difficult tasks, in such domains as autonomous driving \cite{toromanoff2020end, chen2019model}, locomotion control \cite{peng2017deeploco, xie2018feedback}, robotics \cite{levine2016end,neunert2020continuous}, continuous control \cite{banerjee2022boosting, banerjee2022improved,banerjee2022optimal}, and multi-agent systems and control \cite{9022871,CHEN2020109081}. A majority of these successful approaches are purely data-driven and leverage trial-and-error to freely explore the search space.
RL methods work well in simulations, but they struggle with real-world data because of the disconnection between simulated setups and the complexities of real world systems. Major RL challenges \cite{dulac2021challenges}, that are consistently addressed in latest research includes sample efficiency 
\cite{mnih2013playing, barth2018distributed}, high dimensional continuous state and action spaces \cite{dulac2015deep,tessler2019action}, safe exploration \cite{garcia2012safe,gu2022review}, multi-objective and well-defined reward function \cite{knox2023reward,booth2023perils}, perfect simulators and learned model \cite{cutler2015real,osinski2020simulation} and  policy transfer from offline pre-training \cite{levine2020offline,yang2021representation}.

When it comes to machine learning, incorporating mathematical physics into the models can lead to more meaningful solutions. This approach, known as physics-informed machine learning, helps neural networks learn from incomplete physics information and imperfect data more efficiently, resulting in faster training times and better generalization. Additionally, it can assist in tackling high dimensionality applications and ensure that the resulting solution is physically sound and follows the underlying physical law \cite{karniadakis2021physics,banerjee2023physicsinformed,hao2022physics}. 
Among the various sub-fields of ML, RL is the natural candidate for incorporating physics information since most RL-based solutions deal with real-world problems and have an explainable physical structure.

Recent research has seen substantial improvement in addressing the RL challenges by incorporating physics information in the training pipeline.
%
For example, PIRL approaches seek to use physics to reduce high-dimensional continuous states with intuitive representations and better simulation. A low-dimensional representation adhering to physical model PDEs is learned in \cite{gokhale2022physq}, while \cite{cao2023physics} uses features from a supervised surrogate model. Learning a good world model is a quicker and safer alternative to training RL agents in the real world. \cite{ramesh2023physics} incorporate physics into the network for better world models, and \cite{xie2016model} utilize a high-level specification robot morphology and physics for rapid model identification. 

A well-defined reward function is crucial for successful reinforcement learning, PIRL approaches also seek to incorporate physical constraints into the design for safe learning and more efficient reward functions. For example, in \cite{korivand2023inertia} the designed reward incorporates IMU sensor data, imbibing inertial constraints, while in \cite{li2023federated} the physics informed reward is designed to satisfy explicit operational targets.
To ensure safe exploration during training and deployment, works such as \cite{yang2023model,zhao2023barrier} learn a data-driven barrier certificate based on physical property-based losses and a set of unsafe state vectors.

There are several lines of PIRL research dedicated to exploring more efficient exploration of the search space and effective policy deployment for real-world systems. Some approaches were developed to improve simulators for sample efficiency and better sim to real transfer \cite{alam2021physics,lowrey2018reinforcement}. Carefully selecting task-specific state representations \cite{jurj2021increasing, han2022physics}, reward functions \cite{cao2023physical_1, cao2023physical_2}, and action spaces \cite{wang2022ensuring,zhao2023barrier} has been shown to improve both the time to convergence and performance. To sum it up, integrating underlying physics about the learning task structure has been found to improve performance and accelerate convergence. 

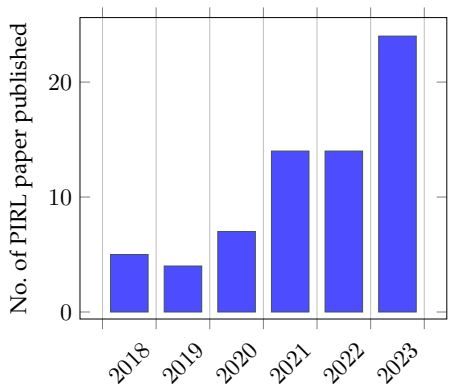
\begin{figure}[h!]
    \centering
    \begin{tikzpicture} [scale=0.9]
    \begin{axis}[
	x tick label style={ rotate = 45,
		/pgf/number format/1000 sep=},
	ylabel= No. of PIRL paper published,
	enlargelimits=0.07,
	ybar interval=0.7,
]
\addplot [cyan!20!black,fill=blue!70!white]
	coordinates {(2023,24) (2022,14)
		 (2021,14) (2020,7) (2019,4) (2018,5) (2017,1) };
\end{axis}
\end{tikzpicture}
    \caption{PIRL papers published over years. This statistic graph the exponential growth of PIRL papers over last six years.}
    \label{fig:timeline}
\end{figure}

Physics-informed Reinforcement Learning (PIRL) has been a growing trend in the literature, as demonstrated in the increasing number of papers published in this area over the past six years, as shown in Figure~\ref{fig:timeline}. The bar chart indicates that this field is gaining more attention, and we can anticipate even more in the future.

\noindent Our contributions in this paper are summarized as follows:
\begin{itemize}
\item[1)] \textit{Taxonomy:} We propose a unified taxonomy to investigate what physics knowledge/processes are modeled, how they are represented, and the strategies to incorporate them into RL approaches.
\item[2)] \textit{Algorithmic Review:} We present state-of-the-art approaches on the physics information guided/ physics informed RL methods, using unified notations, simplified functional diagrams and discussion on latest literature.
\item[3)] \textit{Training and evaluation benchmark Review:} We analyze the evaluation benchmarks used in the reviewed literature, thus presenting popular evaluation and benchmark platforms/ suites for understanding the popular trend and also for easy reference.
\item[4)] \textit{Analysis:} We delve deep into a wide range of model based and model free RL applications over diverse domains. We analyze in detail how physics information is integrated into specific RL approaches, what physical processes have been modeled and incorporated, and what network architectures or network augmentations have been utilized to incorporate physics.
\item[5)] \textit{Open Problems:} We summarize our perspectives on the challenges, open research questions and directions for future research.   
\end{itemize}

\noindent
\textbf{Difference to other survey papers}:\newline
George et al. \cite{karniadakis2021physics} provided one of the most comprehensive reviews on machine learning (ML) in the context of physics-informed (PI) methods, but approaches in the RL domain has not been discussed. The work by Hao et al. \cite{hao2022physics} also provided an overview of physics-informed machine learning, where the authors briefly touch upon the topic of PIRL. Another recent study by Eesserer et al. \cite{eesserer2022guided} showcased the use of prior knowledge to guide reinforcement learning (RL) algorithms, specific to robotic applications. The authors categorize knowledge into three types: expert knowledge, world knowledge, and scientific knowledge. 
Our paper offers a focused and comprehensive review specially on the RL approaches that utilize the structure, properties, or constraints unique to the underlying physics of a process/system. 
Our scope of application domains is not limited to robotics, but also spanning to motion control, molecular structure optimization, safe exploration, and robot manipulation.

The rest of this paper is organized as follows. 
In Section~\ref{sec:PIML_overview}, we provide a brief overview of the Physics informed ML paradigm. In Section~\ref{sec:PIRL_overview_taxonomy-example}, we present RL fundamentals/ framework in ~\ref{sec:RL_funda} and provide a definition with an intuitive introduction to PIRL in ~\ref{sec:PIRL_intro}. Most importantly we introduce a comprehensive taxonomy in ~\ref{Sec: PIRL_taxonomy} threading together physics information types, PIRL methods that implement those information and RL pipeline as a backbone. Later in ~\ref{sec:further_categories} we present and elaborate on two additional categories: Learning architecture and Bias, through which the implementation side of the literature is explained more precisely.
In Section~\ref{sec:PIRL_review,analysis} we present an elaborate review and analysis of latest PIRL literature. In Section~\ref{sec:challenges}, we discuss the different open problems, challenges and research directions that may be addresses in future works  by interested researchers.
Finally Section~\ref{sec:conclusion} concludes the paper.

\begin{table}[t] 
\centering
\caption{A list of abbreviations used in this article.}
\begin{tabular}{|l|l|}
\hline
\multicolumn{2}{|c|}{Abbreviations}\\
\hline
&\\
FSA & Finite State Automata\\
FEA & Finite Element Analysis \\
CFD & Computational Fluid Dynamics \\
MDP & Markov Decision Process\\
MBRL & Model based Reinforcement Learning \\
MFRL & Model Free Reinforcement Learning \\
CBF & Control Barrier Function\\
CBC & Control Barrier Certificate\\
NBC & Neural Barrier Certificate \\
CLBF & Control Lyapunov Barrier Function\\
NBC & Neural Barrier Certificate\\
DFT & Density Functional Theory\\
AC & Actor Critic\\
MPC & Model Predictive Control\\
DDP & Differential Dynamic Programming\\
NPG & Natural Policy Gradient\\
TL & Temporal Logic\\
DMP & Dynamic Movement Primitive \\
WBTG & Whole Body Trajectory Generator \\
DPG & Deterministic Policy Gradient \\
DPPO & Distributed proximal Policy optimization\\
ABM & Adjoint based method \\
APG & Analytic Policy Gradient\\
WBIC & Whole Body Impulse Controller \\
LNN & Lagrangian Neural Network \\
&\\
\hline
\hline
\end{tabular}
\label{table:Abbreviation}
\end{table}

\section{Physics-informed Machine Learning (PIML): an overview}\label{sec:PIML_overview}

The aim of PIML is to merge mathematical physics models and observational data seamlessly in the learning process. This helps to guide the process towards finding a physically consistent solution even in complex scenarios that are partially observed, uncertain, and high-dimensional \cite{kashinath2021physics,hao2022physics,cuomo2022scientific}.
Adding physics knowledge to machine learning models has numerous benefits, as discussed in \cite{kashinath2021physics,meng2022physics}. This information captures the vital physical principles of the process being modeled and brings following advantages

\begin{enumerate}
    \item Ensures that the ML model is consistent both physically and scientifically. 
    \item Increases data efficiency in model training, meaning that the model can be trained with fewer data inputs. 
    \item Accelerates the model training process, allowing models to converge faster to an optimal solution. 
    \item Increases the generalizability of trained models, enabling them to make better predictions for scenarios that were not seen during the training phase. 
    \item Enhances the transparency and interpretability of models, making them more trustworthy and explainable.
\end{enumerate}




\noindent According to literature, there are three strategies for integrating physics knowledge or priors into machine learning models: observational bias, learning bias, and inductive bias.

\textbf{Observational bias:} This approach uses multi-modal data that reflects the physical principles governing their generation \cite{lu2021learning,kashefi2021point,li2020fourier,yang2019conditional}. The deep neural network (DNN) is trained directly on observed data, with the goal of capturing the underlying physical process. The training data can come from various sources such as direct observations, simulation or physical equation-generated data, maps, or extracted physics data induction.

\textbf{Learning bias:} One way to reinforce prior knowledge of physics is through soft penalty constraints. This approach involves adding extra terms to the loss function that are based on the physics of the process, such as momentum or conservation of mass. An example of this is physics-informed neural networks (PINN), which combine information from measurements and partial differential equations (PDEs) by embedding the PDEs into the neural network's loss function using automatic differentiation \cite{karniadakis2021physics}. Some prominent examples of soft penalty based approaches includes statistically constrained GAN \cite{wu2020enforcing}, physics-informed auto-encoders \cite{erichson2019physics} and encoding invariances by soft constraints in the loss function InvNet \cite{shah2019encoding}. 

\textbf{Inductive biases:} Custom neural network-induced 'hard' constraints can incorporate prior knowledge into models. For instance, Hamiltonian NN \cite{greydanus2019hamiltonian} draws inspiration from Hamiltonian mechanics and trains models to respect exact conservation laws, resulting in better inductive biases. Lagrangian Neural Networks (LNNs) \cite{cranmer2020lagrangian} introduced by Cranmer et al. can parameterize arbitrary Lagrangians using neural networks, even when canonical momenta are unknown or difficult to compute. Meng et al. \cite{meng2022learning} uses a Bayesian framework to learn functional priors from data and physics with a PI-GAN, followed by estimating the posterior PI-GAN's latent space using the Hamiltonian Monte Carlo (HMC) method. Additionally, DeepONets \cite{lu2021learning} networks are used in PDE agnostic physical problems.

\section{Physics-Informed Reinforcement Learning: Fundamentals, Taxonomy and Examples} \label{sec:PIRL_overview_taxonomy-example}
In this section, we will explain how physics information can be integrated into reinforcement learning applications. 

\subsection{RL fundamentals}\label{sec:RL_funda}

\begin{figure}[h!]
    \centering\includegraphics[width=0.8\columnwidth]{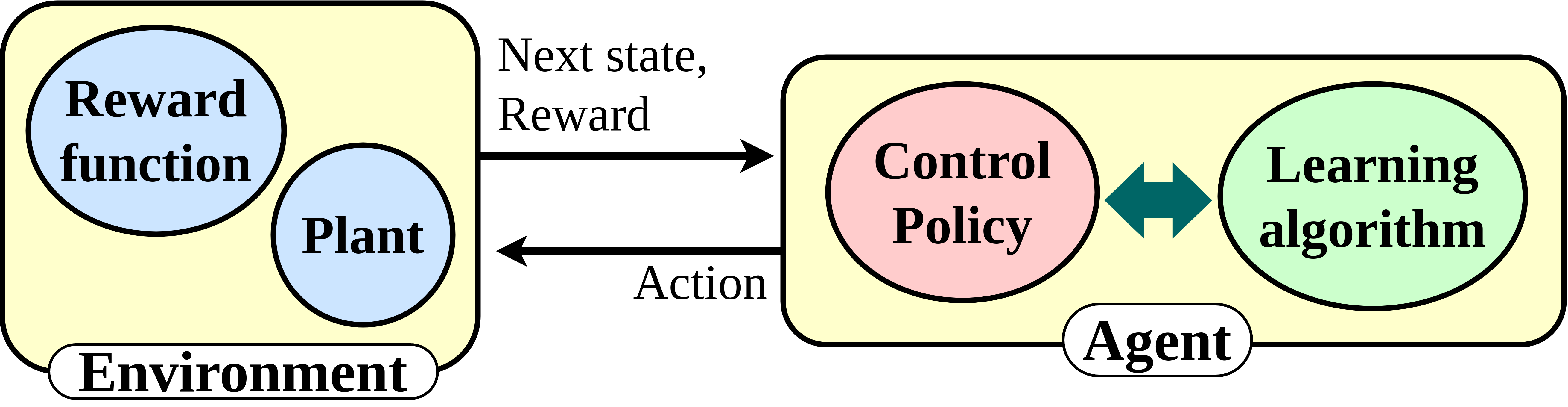}
    \caption{ Agent-environment framework, of RL paradigm. Here the reward generating function and the system/ plant is abstracted as the environment. And the control policy (e.g. a DNN) and the learning algorithm, forms the RL agent.}
    \label{fig:RL_basic}
\end{figure}

RL algorithms use reward signals from the environment to learn the best strategy to solve a task through trial and error. They effectively solve sequential decision-making problems that follow the Markov Decision Process (MDP) framework. 
In the RL paradigm, there are two main players: the agent and the environment. The environment refers to the world where the agent resides and interacts. Through agent-environment interactions the agent perceives the state of the world and decides on the appropriate action to take.

The agent-environment RL framework, see Fig.~\ref{fig:RL_basic}, is a large abstraction of the problem of goal-directed learning from interaction \cite{sutton2018reinforcement}. The details of control apparatus, sensors, and memory are abstracted into three signals between the agent and the environment: the control/ action, the state and the reward.
Though typically, the agents computes the rewards, but by the current convention anything that cannot be changed arbitrarily by the agent is considered outside of it and hence the reward function is shown as a part of the environment.

MDP is typically represented by the tuple $(\mathcal{S},\mathcal{A}, R, P, \gamma)$, where $\mathcal{S}$  represents the states of the environment, $\mathcal{A}$ represents set of actions that the RL agent can take. Reward function may be typically represented as $R(s_{t+1},a_t)$ a function of next state and current action. The function generates the reward due to action induced state transition from $s_t$ to $s_{t+1}$. $P(s_{t+1}|s_t,a_t)$ is the environment model that returns the probability of transitioning to state $s_{t+1}$ from $s_t$. Finally the discount factor $\gamma \in [0,1]$, determines the amount of emphasis given to the immediate rewards relative to that of future rewards.

\begin{figure}[t]
\centering
\subf{\includegraphics[width=0.4\columnwidth]{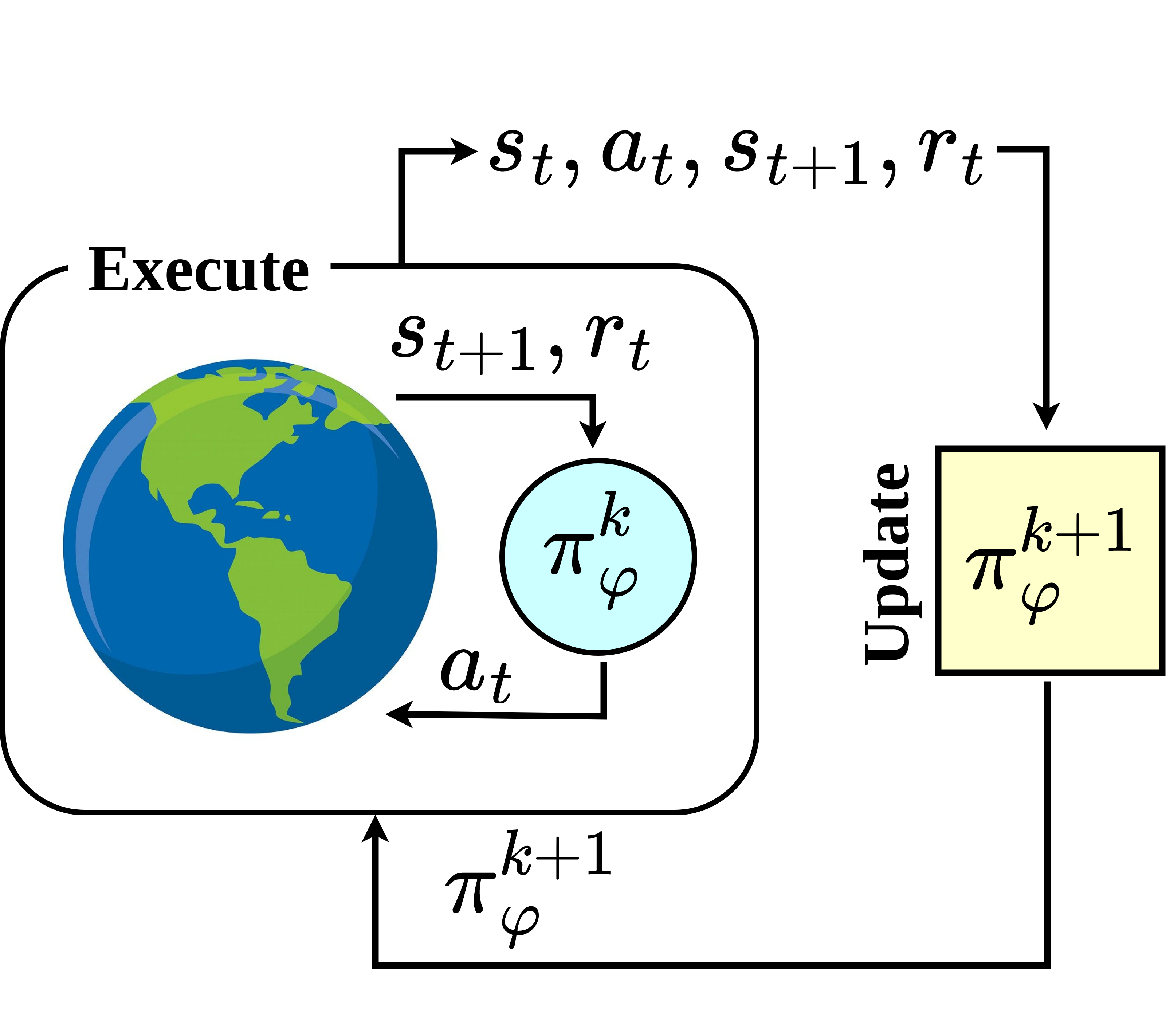}}
     {(a) Online RL (model-free) }
     \label{fig:online_RL}

\subf{\includegraphics[width=0.4\columnwidth]{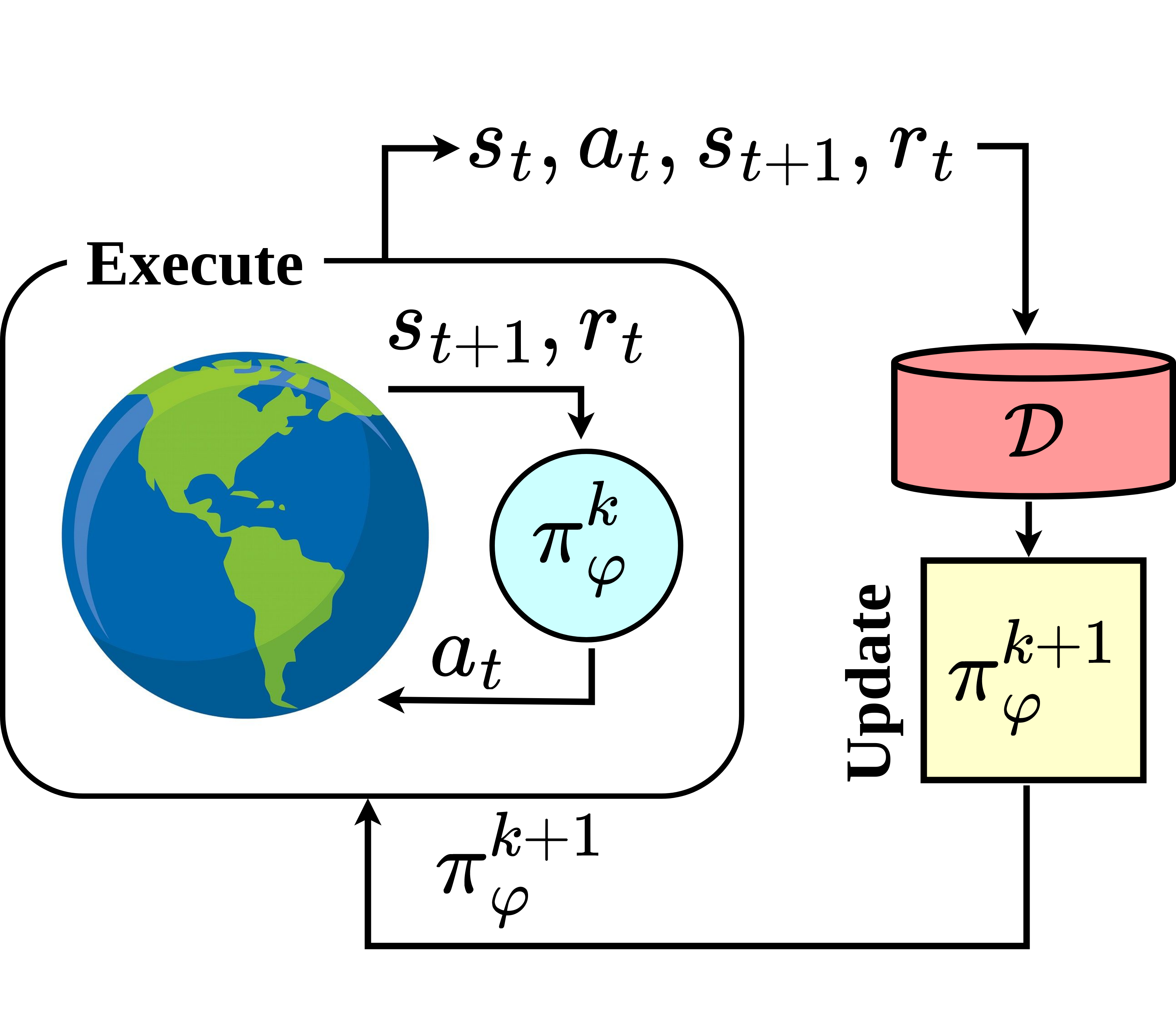}}
     {(b) Off-policy RL (model-free) }
     \label{fig:off_policy_RL_MF}

\subf{\includegraphics[width=0.6\columnwidth]{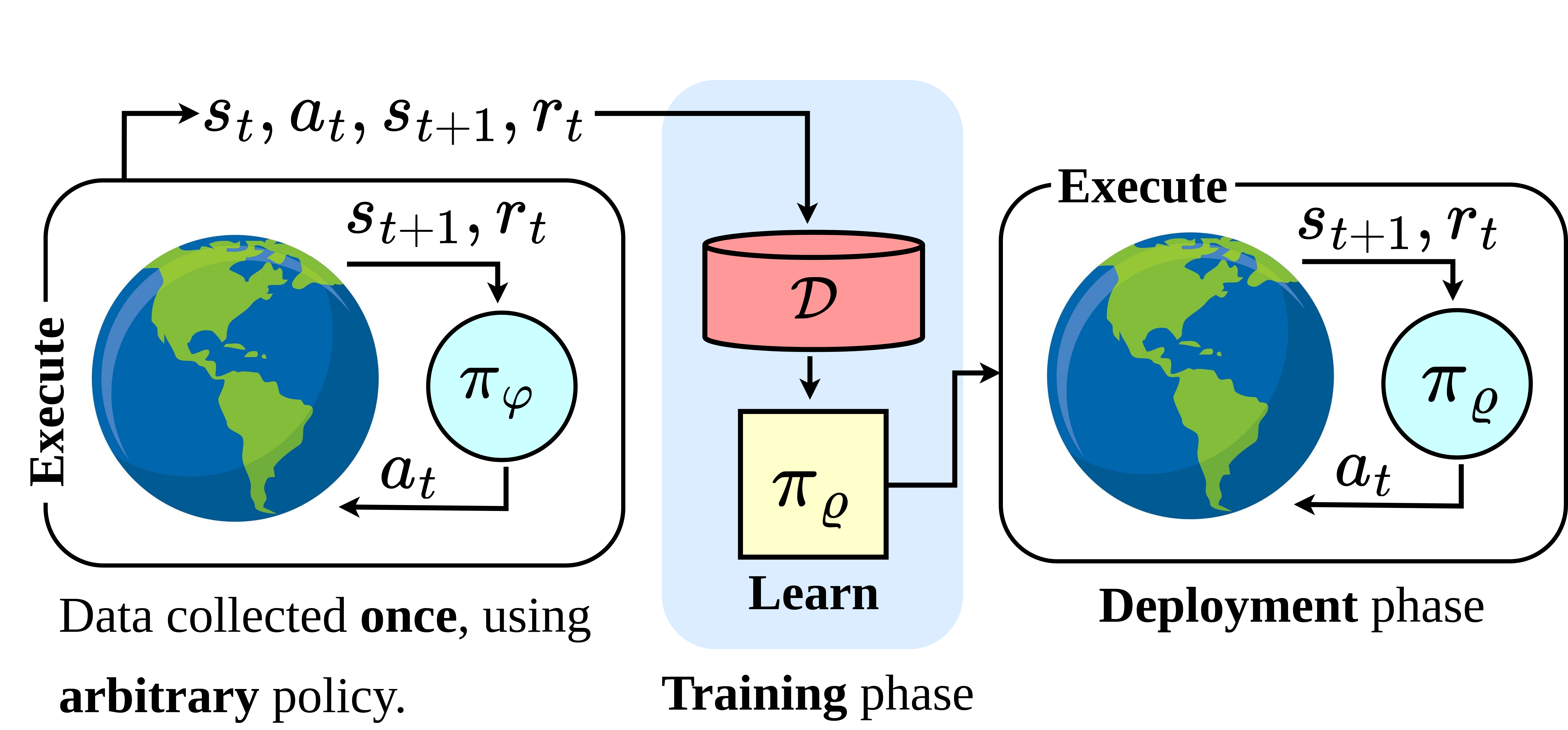}}
     {(c) Offline RL  }
     \label{fig:offline_RL}

\subf{\includegraphics[width=0.6\columnwidth]{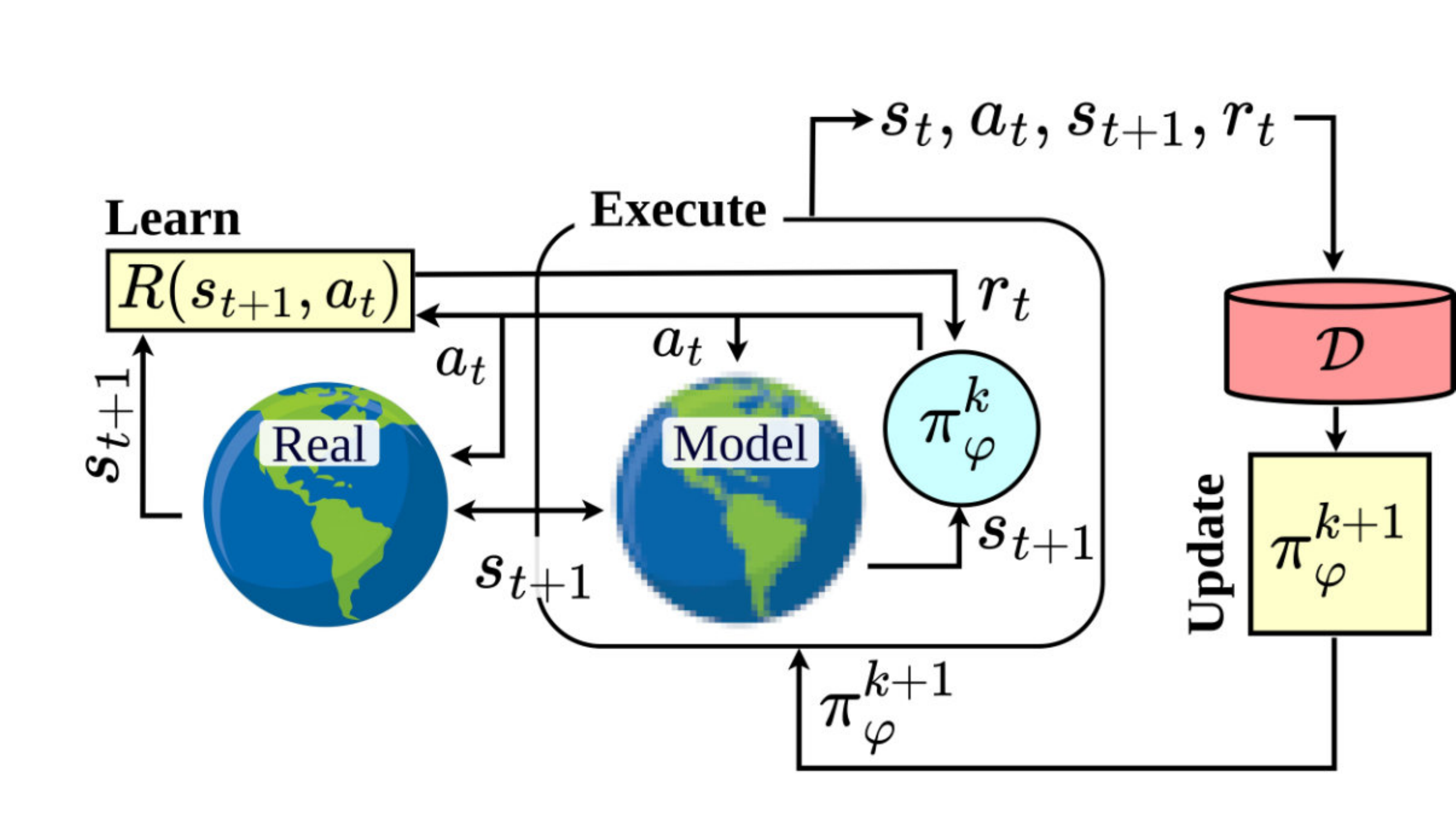}}
     {(d) Off-policy RL (model-based)}
     \label{fig:off_policy_Rl_MB}

\caption{Typical RL architectures, based on model use and interaction with the environment.}
\label{fig:Typical RL arch.}
\end{figure}

The RL framework typically organizes the agent's interactions with the environment into episodes. In each episode, the agent starts at a particular initial state $s_1$ sampled from an initial distribution $p(s_1)$, which is part of the state space $\mathcal{S}$ of the MDP. At each timestep $t$, the agent observes the current state $s_t \in \mathcal{S}$ and samples an action $a_t \in \mathcal{A}$ from its latest policy $\pi_{\phi}(a_t | s_t)$ based on the state $s_t$, where $\phi$ represents the policy parameters. The action space of the MDP is denoted by $\mathcal{A}$.

Next, the agent applies the action $a_t$ into the environment, which results in a new state $s_{t+1}$ given by the dynamics of the MDP, i.e., $s_{t+1} \sim p(s_{t+1}| s_t, a_t)$. The agent also receives a reward $r_t = R(s_{t+1}, a_t)$, which can be construed as the desirability of a certain state transition from the context of the given task. 
The above process is repeated up to a certain time horizon $T$, which may also be infinite. The agent-environment interaction is recorded as a trajectory, and the closed-loop trajectory distribution for the episode $t=1,\cdots, T$ can be represented by, 
\begin{align}
   p_{\phi}(\tau) = &{p_{\phi}(s_1,a_1,s_2,a_2,\cdots,s_T, a_{T},s_{T+1})} \\
  = & p(s_1)\prod_{t=1}^T 
     \pi_{\phi}(a_t|s_t) p(s_{t+1}|s_t,a_t), 
\end{align}
where $\tau =(s_1,a_1,s_2,a_2,\cdots,s_T, a_{T},s_{T+1})$ represents the sequence of states and control actions. The objective is to find an optimal policy represented by the parameter,
 \begin{align}
     \phi^* = \text{arg max}_{\phi}\, \underbrace{{\mathbf E}_{\tau \sim p_{\phi}(\tau)}\, \Big[\sum_{t=1}^T\,\gamma ^{t} R( a_t,s_{t+1})\Big]}_{\mathcal{J}(\phi)},
     \end{align}
which maximizes the objective function $\mathcal{J}(\phi)$, $\gamma$ is a parameter called discount factor, where $0\leq \gamma \leq 1$. $\gamma$ determines the present value of the future reward, i.e., a reward received at $k$ timesteps in the future is worth only $\gamma ^{k-1}$ times what it would be worth if received immediately. 

\vspace{6px}
\noindent \textbf{Model-free and model-based RL:}
In RL, algorithms can be classified based on whether the environment model is available during policy optimization. The environment dynamics are represented as $p(s_{t+1}, r_{t}) = Pr(s_{t+1}, r_{t}|s_{t}, a_{t})$, which means that given a state and action, the environment model can predict the state transition and the corresponding reward. Access to an environment model allows the agent to plan and choose between options and also improves sample efficiency compared to model-free approaches. However, the downside is that the environment's groundtruth model is typically not available, and learning a perfect model of the real world is challenging. Additionally, any bias in the learned model can lead to good performance in the learned model but poor performance in the real environment.

\vspace{6px}
\noindent \textbf{Online, Off-policy and Offline RL:}
Online RL algorithms, e.g. PPO, TRPO, and A3C, optimize policies by using only data collected while following the latest policy, creating an approximator for the state or action value functions, used to update the policy. Off-policy RL algorithms, e.g. SAC, TD3 and IPNS, involve the agent updating its policy and other networks using data collected at any point during training. This data is stored in a buffer called the experience replay buffer and is in the form of tuples. Mini-batches are sampled from the buffer and used for the training process. Offline RL algorithms use a fixed dataset called $\mathcal{D}$ collected by a policy $\pi_{\zeta}$ to learn the optimal policy. This allows for the use of large datasets collected previously.

Combining model-free/model-based with online/off-policy/offline categorization, typical RL architectures can be presented as Fig.~\ref{fig:Typical RL arch.}.

\begin{figure*}[h!]
    \centering\includegraphics[width=0.8\textwidth]{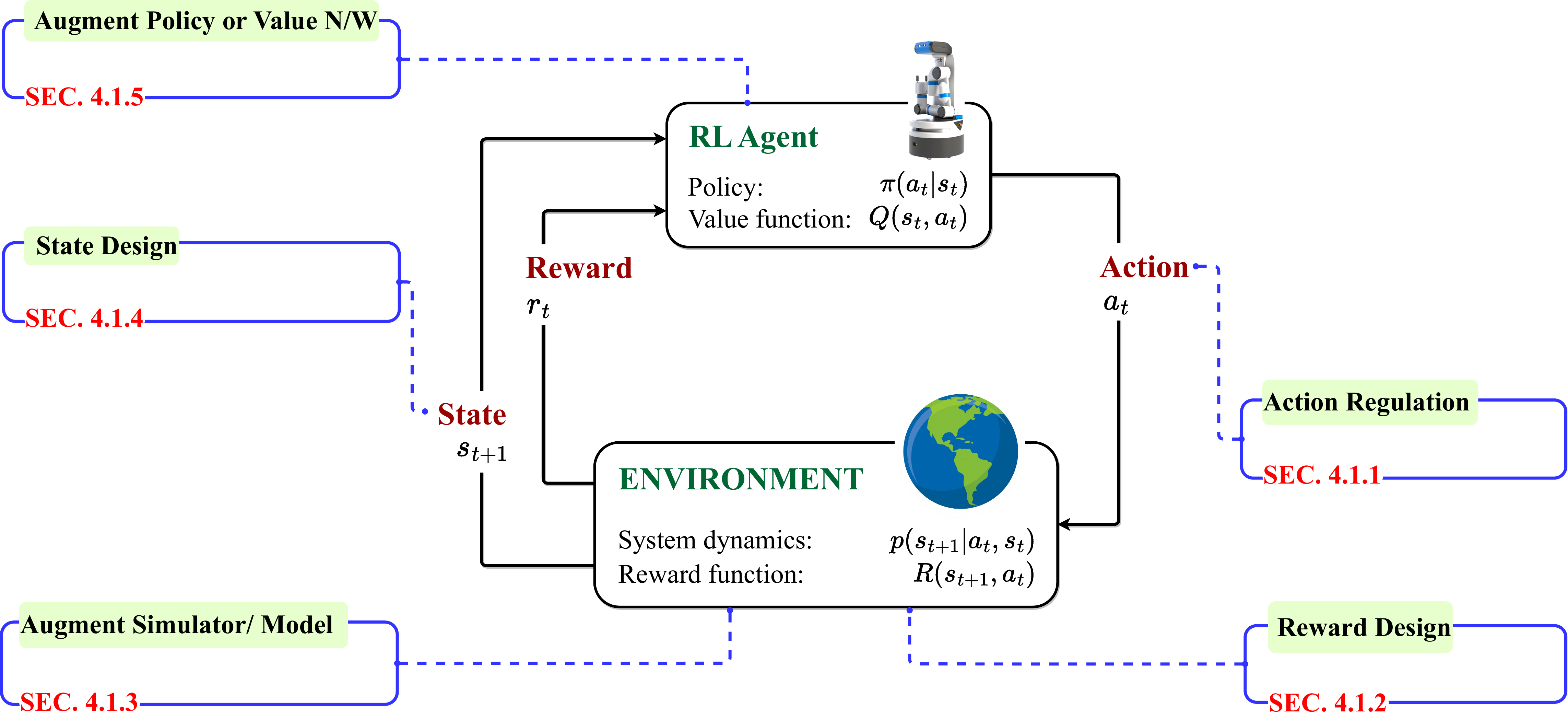}
    \caption{Map of physics incorporation (PI) in the conventional Reinforcement Learning (RL) framework.}
    \label{fig:PI_aug_RL}
\end{figure*}

\subsection{PIRL: Introduction}\label{sec:PIRL_intro}
\subsubsection{Definition}
The concept of physics-informed RL involves incorporating physics structures, priors, and real-world physical variables into the policy learning or optimization process. Physics induction helps improve the effectiveness, sample efficiency and accelerated training of RL algorithms/ approaches, for complex problem-solving and real-world deployment. Depending on the specific problem or scenario, different physics priors can be integrated using various RL methods at different stages of the RL framework, see Fig.~\ref{fig:PI_aug_RL}. 

\subsubsection{Intuitive introduction to physics priors in RL}

\begin{figure}[h!]
    \centering\includegraphics[width=0.6\linewidth]{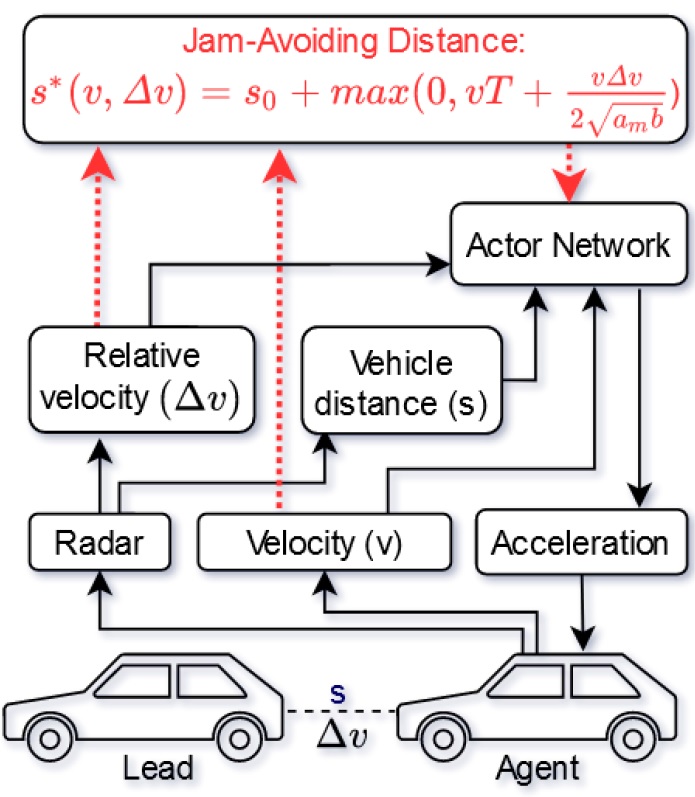}
    \caption{ An illustrative example of physics incorporation in RL application, \cite{jurj2021increasing}. Here the RL agent is fed with an additional state variable: jam avoiding distance, which is based on desired physical parameters and primary state variables.}
    \label{fig:first_phy_example}
\end{figure}

Physics priors come in different forms, like intuitive physical rules or constrains, underlying mathematical/ guiding equations and physics simulators, to name a few. Here we discuss a couple of intuitive examples.
In \cite{xie2016model}, the physical characteristics of the system were utilized as priors. The high-level specifications of a robot's morphology such as the number and connectivity structure of links were used as physics priors. This feature based representation of the system dynamics enabled rapid model identification in this model based RL setup. In another example, pertaining to adaptive cruise control problem, \cite{jurj2021increasing} (see Fig.\ref{fig:first_phy_example}), physics information in the form of \say{jam-avoiding distance} (based on desired physical parameters e.g. velocity and acceleration constraints, minimum jam avoiding distance etc.) is included in state space input to the RL agent. Physics info. incorporation results in a RL controller which performs with less collisions and enables more equidistant travel.

\subsection{PIRL Taxonomy}\label{Sec: PIRL_taxonomy}
\subsubsection{Physics information (types): representation of physics priors} 
There are different types/ forms of physics information, e.g. mathematical representation of the physical system like PDE/ODE and physics enriched simulators.
Based on the type of the physics information representation, works can be typically categorized as follows. 

\begin{enumerate} 
   \item[1)] \textit{Differential and algebraic equations (DAE):} Many works use system dynamics representations, such as partial/ordinary differential equations (PDE/ ODE) and boundary conditions (BC), as physics priors primarily through PINN and other special networks. For example in transient voltage control application \cite{gao2022transient}, a PINN is trained using PDE of transient process. The PINN learns a physical constraint which it transfers to the loss term of the RL algorithm.
   \item[2)] \textit{Barrier certificate and physical constraints (BPC):} It is imperative to regulate agent exploration in safety-critical applications of reinforcement learning. One way it is addressed in recent research is through the use of optimization-based control theoretic constraints. Use of concepts like control Lyapunov function (CLF)\cite{li2019temporal, choi2020reinforcement}, barrier certificate/ barrier function (BF), control barrier function/ certificate (CBF/ CBC) \cite{cheng2019end,cai2021safe} is made in recent safety critical RL applications.
   Barrier certificate is generally used to establish a safe set of desired states for a system. A control barrier function is then employed to devise a control law that keeps the states within the safety set.
   In certain scenarios barrier functions are represented as NNs and learned through data driven approaches \cite{zhao2023barrier,zhao2022safe}.  
   In above control theoretic approaches the system dynamics either partial or learnable and safety sets represent the primary physical information.
   For more details on CBFs refer \cite{ames2019control}. Additionally safety in the learning process may also be ensured by incorporating physical constraints into the RL loss \cite{li2021safe,chen2022physics}.

   \item[3)] \textit{Physics parameters, primitives and physical variables (PPV):} Physics values extracted/ derived from the environment or system has been directly used by RL agents in form of physics parameters \cite{siekmann2021sim}, dynamic movement (physics) primitives \cite{bahl2021hierarchical}, physical state \cite{jurj2021increasing} and physical target \cite{li2023federated}. For example in \cite{li2023federated}, the reward is created to meet two physical objectives/ targets: operation cost and self-energy sustainability. In an adaptive cruise control problem \cite{jurj2021increasing}, authors use desired physical parameters e.g. velocity and acceleration constraints and minimum jam avoiding distance, as a state space input.

   \item[4)] \textit{Offline data and representation (ODR):}
   For the improvement simulator based training, especially during sim-to-real transfer,  non-task- specific-policy data collected from real robot has been used to train RL agents in offline setting along with simulators \cite{golemo2018sim} and as hardware data to seed simulators \cite{lowrey2018reinforcement}.

    Another popular way of extracting physics information from environment is learning physically relevant low dimensional representation from observations \cite{gokhale2022physq,cao2023physics}.
    For example, in \cite{gokhale2022physq}, PINN is used to extract physically relevant information about the hidden state of the system, which is further used to learn a Q-function for policy optimization.
    
   \item[5)] \textit{Physics simulator and model (PS): }
   Simulators provide a easy way of experimenting with RL algorithms without exposing the agent e.g. a robot to the wear and tear of the real environment. 

   Apart from serving as test-beds for RL algorithms, simulators are also used alongside RL algorithms to impart physical correctness or physics awareness in the data or training process.
   For example in order to improve motion capture and imitation of given motion clips, \cite{chentanez2018physics} have used rigid body physics simulations to solve the rigid body poses closely following the motion capture clip. In \cite{garcia2020physics}, using a physics simulator, a residual agent is able to learn how to improve user input in order to achieve a task while staying true to the original input and expert-recorded trajectories.
   
   In the MBRL setting the system model can be: 1) completely known, 2)
   partially known or 3) completely unknown. RL algorithms typically addresses the last two types, since it deals with environments whose dynamics is complex and difficult to ascertain through classical approaches. In such cases a DNN based data-driven approach is generally utilized to learn the system model completely or enrich the existing partial or basic model of the environment. 
   In \cite{han2022physics} a data driven surrogate traffic flow model is learned that generates synthetic data. This data is later used by the agent in an offline learning process, followed by an online control process.
   In \cite{ramesh2023physics} learns environment and reward models by using Lagrangian NNs \cite{cranmer2020lagrangian}. LNNs are models that are able to Lagrangian functions straight from data gathered from agent-environment interactions.

   \item[6)] \textit{Physical properties (PPR):} 
   Fundamental knowledge regarding the physical structure or properties pertaining to a system has been used in a number of works. For example system morphology, system symmetry \cite{huang2023symmetry}
   
\end{enumerate}

\subsubsection{PIRL methods: physics prior augmentations to RL} 
PIRL methods highlights and discusses about the different components of the typical RL paradigm e.g. state space, action space, reward function and agent networks (policy and value function N/W), that has been directly modified/ augmented through the incorporation of physics information.
\begin{itemize}
    \item[1)] \textit{State design:} This category is concerned with the observed state space of the environment or model. The PIRL approaches, typically modifies or expands the state representation in order to make it more instructive. 
    Works include state fusion using additional information from environment \cite{jurj2021increasing} and other agents \cite{shi2023physics}, state as extracted features from robust representation \cite{cao2023physics}, learned surrogate model generated data as state \cite{han2022physics} and state constraints \cite{zhang2022barrier}.
    \item[2)] \textit{Action regulation:} This pertains to modifying the action value, which is often achieved through PIRL approaches that impose constraints on the action value to ensure safety protocols are implemented \cite{li2021safe,cheng2019end}. 
    \item[3)] \textit{Reward design:} It concerns approaches that induce physics information through effective reward design or augmentation of existing reward functions with bonuses or penalties \cite{dang2022towards,luo2020kinematics}.
    \item[4)] \textit{Augment policy or value N/W:} These PIRL approaches incorporate physics principles via methods like, adjusting the update rules and losses of the policy \cite{bahl2020neural,margolis2021learning}, value functions \cite{mukherjee2023bridging,park2023physics} and making direct changes to their underlying network structure \cite{cao2023physical_2}. 
    Works with novel physics based losses \cite{mora2021pods,xu2022accelerated} and constraints for policy or value function learning \cite{gao2022transient} are also included. 

    \item[5)] \textit{Augment simulator or model: } This category encompasses those works that develops improved simulators through incorporation of underlying physics knowledge thereby allowing for more accurate simulation of real-world environments. 
    Works include physics based augmentation of DNN based learnable models for accurate system model learning \cite{lee2020context,ramesh2023physics}, improved simulators for sim-to-real transfer \cite{golemo2018sim, lowrey2018reinforcement} and physics informed learning for partially known environment model \cite{liu2021physics}. 
    %
    
\end{itemize}

\begin{figure*}[t]
  \centering
  \includegraphics[width=0.8\linewidth]{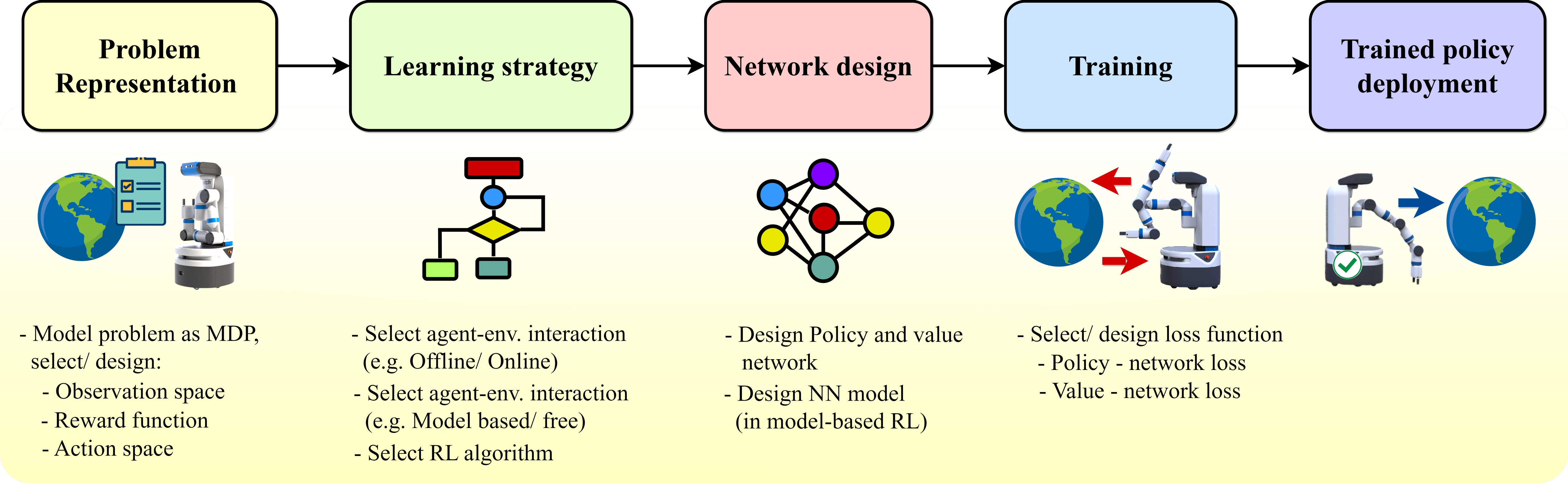}
  \caption{Deep Reinforcement Learning Pipeline. Here the problem is first modeled as a MDP, clearly defining the state, action and reward spaces. Followed by selecting the RL algorithm as Learning strategy and then selecting/ designing the policy and/or value networks in network design stage. Finally the agent is trained using default/ custom loss function in training stage and  finally deployed. }
  \label{fig:RL_pipeline_steps}
\end{figure*}

\subsubsection{RL Pipeline}
A typical RL pipeline can be represented into four functional stages namely, the problem representation, learning strategy, network design, training and trained policy deployment. These stages are elaborated as follows:
\begin{itemize}
    \item [1.] \textit{Problem Representation:} In this stage,  a real-world problem is modeled as a Markov Decision Process (MDP) and thereby described using formal RL terms. The main challenge is to choose the right observation vector, define the reward function, and specify the action space for the RL agent so that it can perform the specified task properly.
    \item [2.] \textit{Learning strategy:} In this stage, the decisions are made regarding the type of agent-environment interaction e.g. in terms of environment model use, learning architecture and the choice of RL algorithm.
    
    \item [3.] \textit{Network design:} Here the finer details of the learning framework are decided and customized where needed. Decisions are made regarding the type of constituent units (e.g. layer types, network depth etc.) of underlying Policy and value function networks.
    
    \item [4.] \textit{Training:} The policy and allied networks are trained in this stage. It also represents training augmentation approaches like Sim-to-real, that helps is reducing discrepancy between simulated and real worlds.
    
    \item [5.] \textit{Trained policy deployment:} At this stage the policy is completely trained and is deployed for solving the concerned task.
\end{itemize}

\begin{figure}[h!]
    \centering\includegraphics[width=0.6\columnwidth]{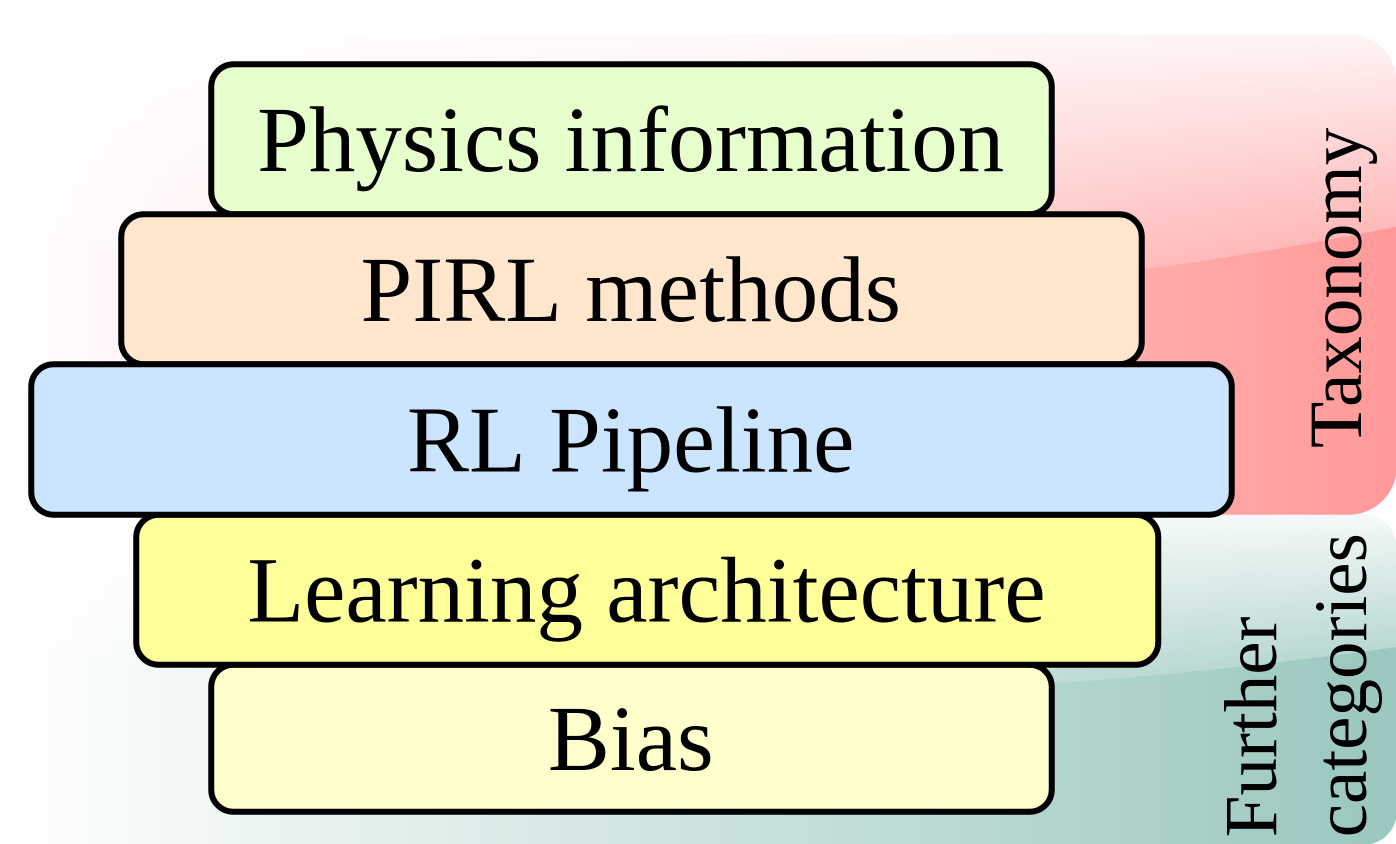}
    \caption{ PIRL taxonomy and further categories. Physics information (types), the RL methods that incorporate them and the underlying RL pipeline constitutes the PIRL-taxonomy, see Fig.~\ref{fig:PIRL_taxonomy}. bias (sec.~\ref{sec:Bias}) and Learning architecture (sec.~\ref{Sec: Learning architecture }) are two additional categories which has been introduced to better explain the implementation of PIRL.}
    \label{fig:PIRL_taxonomy_and_further_categories}
\end{figure}

\begin{figure*}[h!]
    \centering\includegraphics[width=0.91\textwidth]{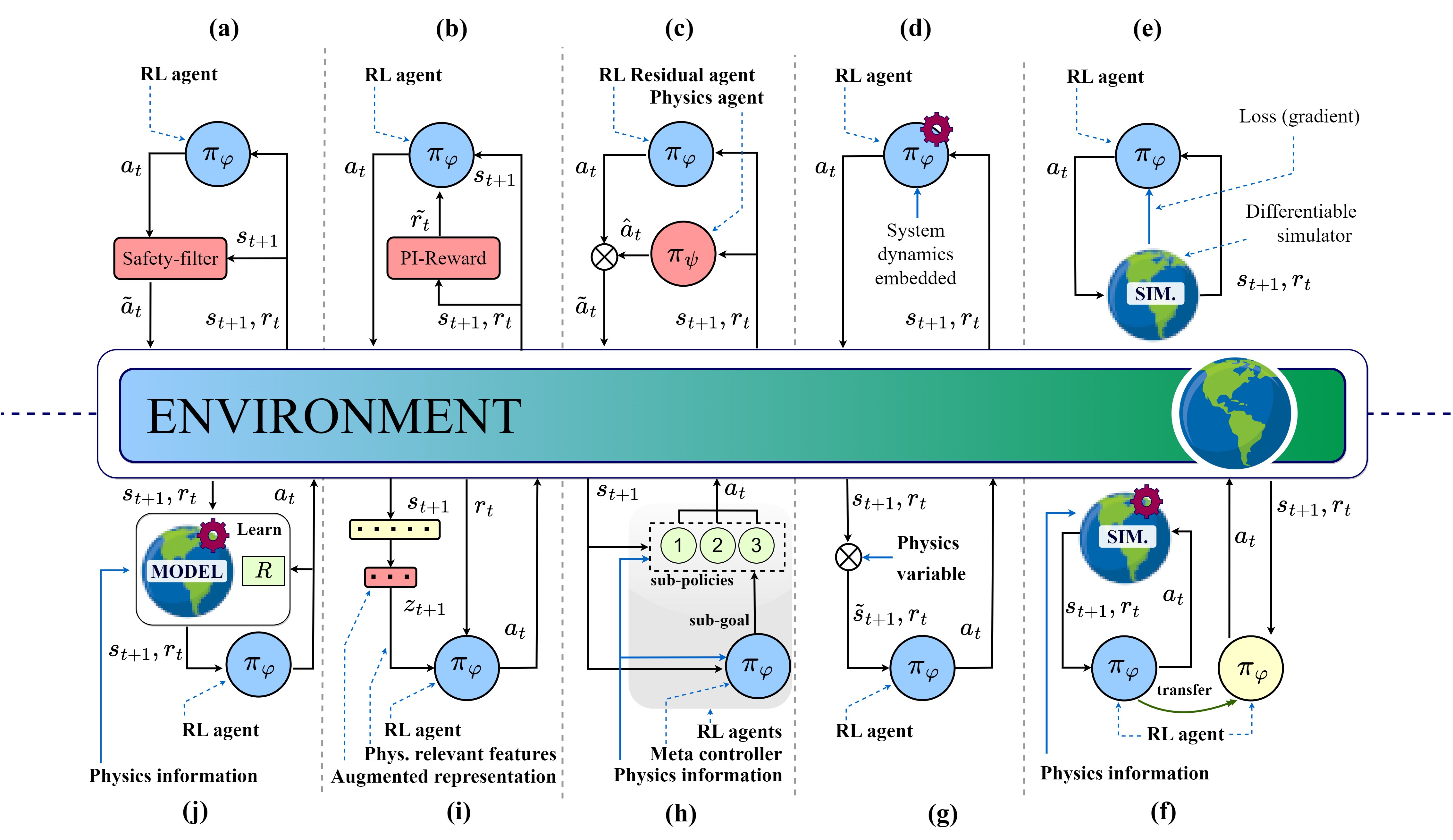}
    \caption{  Typical RL architectures with physics information incorporation (a) Safety filter (b) PI-reward (c) Residual agent (d) Physics embedded network (e) Differentiable simulator (f) Sim-to-Real (g) Physics variable (h) Hierarchical RL (i) Data augmentation (j) PI model identification. To keep illustrations simple, we have not included ancillary networks e.g. value function networks above.}
    \label{fig:PIRL_learning architectures}
\end{figure*}

\subsection{Further categorization} \label{sec:further_categories}
In this section we introduce a couple of additional categorizations: Bias and Learning architecture. These categories are not part of the taxonomy that we have discussed in the previous section, see Fig.\ref{fig:PIRL_taxonomy_and_further_categories}. They provide an additional perspective to the PIRL approaches presented here.

\subsubsection{Bias} \label{sec:Bias}
PI approaches in ML paradigm, mentions of different kind of biases or categories of methods of physics incorporation in ML models. In order to relate to that existing taxonomy used in PIML methods, in Table~\ref{table:PIRL_Characteristic_MF} and Table~\ref{table:PIRL_Characteristics_MB}., we include corresponding bias categories to each of the PIRL entries.

\subsubsection{Learning architecture}\label{Sec: Learning architecture }
We also categorize PIRL algorithms based on the alterations that they introduce to the conventional RL learning architecture to incorporate physics information/ priors. As listed and discussed below they help us understand the PIRL methods from an architectural point of view. In the literature review section we use the aid of such learning architecture categories to group and discuss the PIRL methods. 

\begin{itemize}
    \item[1)] \textit{Safety filter: } This category includes approaches that has a PI based module which regulates the agent's exploration ensuring safety constraints, for reference see Fig. \ref{fig:PIRL_learning architectures}(a). In this typical architecture the safety-filter module takes action $a_t$ from RL agent $\pi_{\varphi}$, and state information $(s_{t})$ and refines the action, giving $\tilde{a}_t$.
    \item[2)] \textit{PI reward:} This category includes approaches where physics information is used to modify the reward function, see Fig.\ref{fig:PIRL_learning architectures}(b) for reference. Here the PI-reward module augments agent's extrinsic reward $(r_t)$ with a physics information based intrinsic component, giving $\tilde{r}_t$.
    \item[3)] \textit{Residual learning:} Residual RL is an architecture which typically consists of two controllers: a human designed controller and a learned policy \cite{johannink2019residual}.  In PIRL setting the architecture consists of a physics informed controller $\pi_{\psi}$ along with the data-driven DNN based policy $\pi_{\varphi}$, called residual RL agent, see Fig. \ref{fig:PIRL_learning architectures}(c).
    \item[4)] \textit{Physics embedded network:} In this category physics information e.g. system dynamics is directly incorporated in the policy or value function networks, see Fig.\ref{fig:PIRL_learning architectures}(d) for reference. 
    \item[5)] \textit{Differentiable simulator:} Here the approaches have use differentiable physics simulators, which are non-conventional/ or adapted simulators and explicitly provides loss gradients of simulation outcome w.r.t. control action, see Fig.\ref{fig:PIRL_learning architectures}(e) for reference.
    \item[6)] \textit{Sim-to-Real:} In Sim-to-real architecture, the agent is first trained on a simulator or source domain and is later transferred to a target domain for deployment. In certain cases the transfer is followed by fine-tuning at the target domain, see Fig.\ref{fig:PIRL_learning architectures}(f) for reference.
    \item[7)] \textit{Physics variable:} This architecture encompasses all those approaches where physical parameters, variables or primitives are introduced to augment components (e.g. states and reward) of the RL framework. For reference see Fig.\ref{fig:PIRL_learning architectures}(g).
    \item[8)] \textit{Hierarchical RL:} This category includes hierarchical and curriculum learning based approaches, Fig.\ref{fig:PIRL_learning architectures}(h) for reference. In a hierarchical RL (HRL) setting a long horizon decision making task is broken into simpler sub-tasks autonomously. In curriculum learning a complex task is solved by learning to solve a series of increasingly difficult tasks.
    In both HRL and CRL physics is typically incorporated into all the policy (including meta and sub-policies) and value networks. Approaches here are mostly extensions of physics-embedded networks (Fig.\ref{fig:PIRL_learning architectures}(d)), as used in non-HRL/ CRL settings.
    \item[9)] \textit{Data augmentation:} This category includes approaches where the input state is replaced with a different or augmented form of it, e.g. low dimensional representation so as to derive special and physically relevant features out of it. See Fig.\ref{fig:PIRL_learning architectures}(i) for reference. In this typical architecture, the state vector $s_{t+1}$ is transformed into an augmented representation $z_{t+1}$. Physically relevant features are then extracted from it and used by the RL agent ($\pi_{\varphi}$).
    \item[10)]\textit{PI model identification: } This architecture represents those PIRL approaches, especially in data-driven MBRL setting where physics information is directly incorporated into the model identification process. For reference see Fig.\ref{fig:PIRL_learning architectures}(j).
\end{itemize}

\section{PIRL: Review and Analysis}\label{sec:PIRL_review,analysis}
In this section we provide a indepth review of latest works in PIRL, followed by a review of the popular datasets. We also include an analysis of the algorithms and their derivatives, and discuss crucial insights.
\subsection{Algorithmic review}
We provide a detailed overview of the PIRL approaches as identified by our literature review in Table~\ref{table:PIRL_Characteristic_MF} and Table~\ref{table:PIRL_Characteristics_MB}. We have structured our discussion according to the methods of the introduced taxonomy (see \ref{Sec: PIRL_taxonomy}) since they form a bridge between the physics information sources and practical applications. We also use learning architecture categories as introduced in ~\ref{Sec: Learning architecture }, to better explain the PIRL methods.

\begin{figure*}[t]
\centering\includegraphics[width=0.9\linewidth]{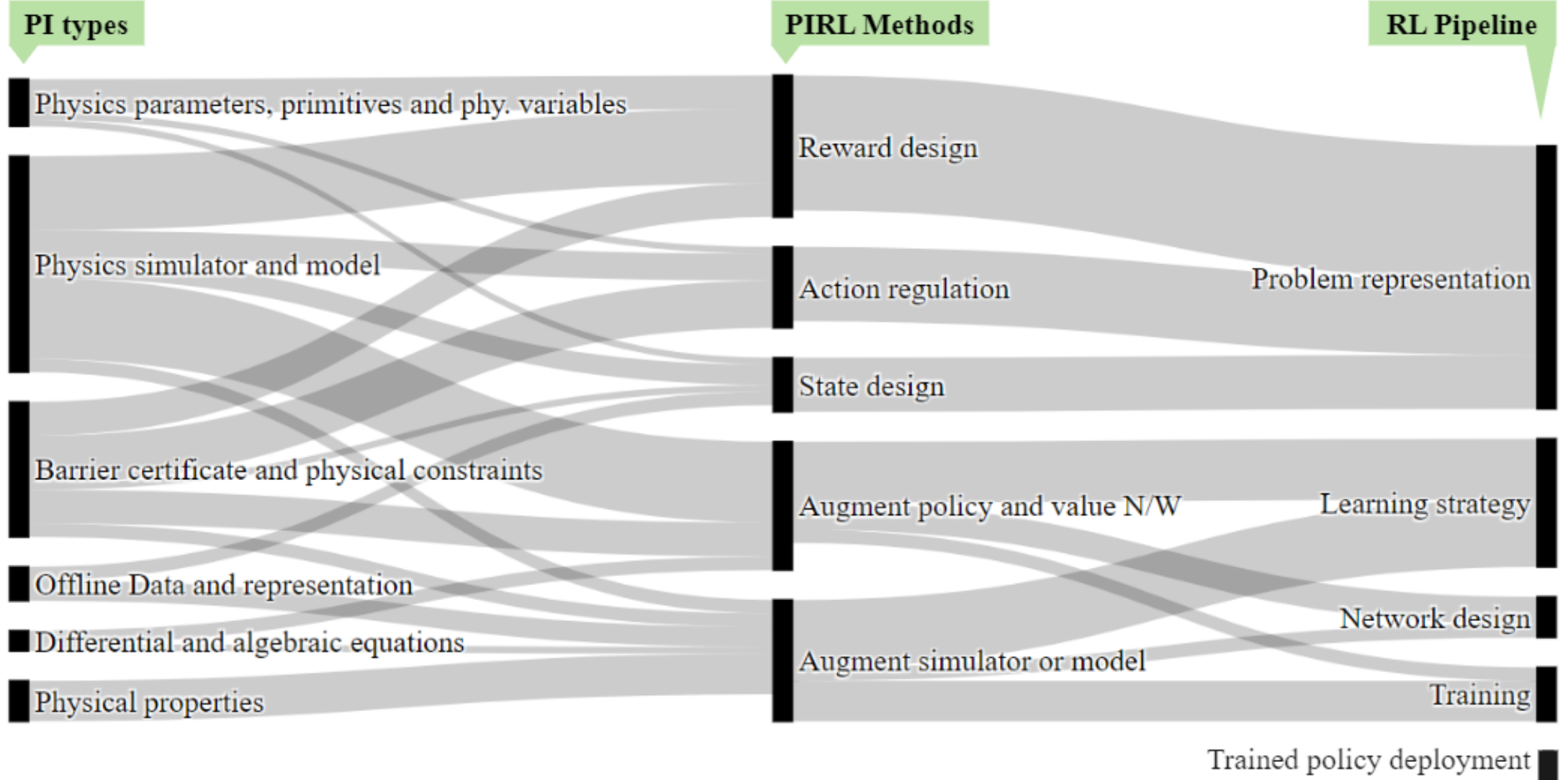}
    \caption{Taxonomy, the diagram connects PI types with PIRL methods and then to the RL pipeline backbone. The connection thickness represents the quantity of work done which corresponds to those components/ categories.}
    \label{fig:PIRL_taxonomy}
\end{figure*}

\subsubsection{State design:}

\begin{figure}[H]
    \centering\includegraphics[width=0.7\linewidth]{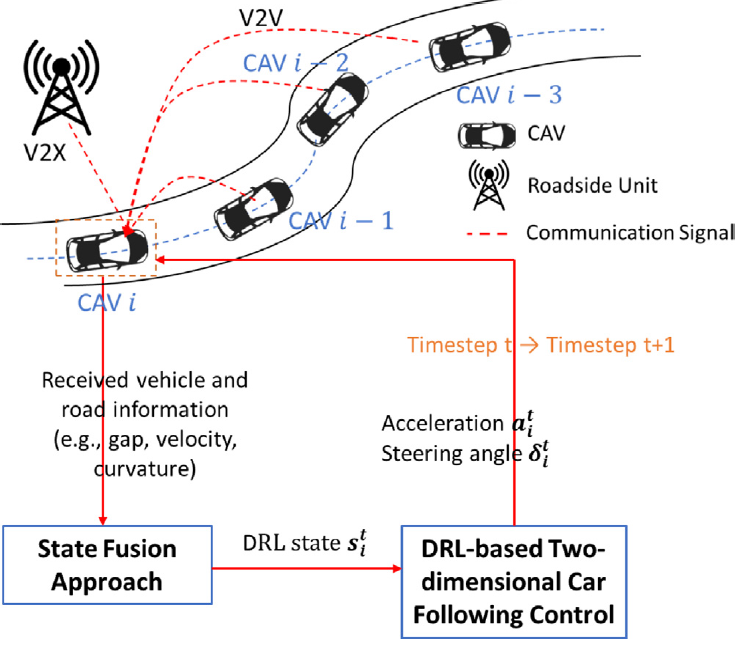}
    \caption{ Example of state design, through physics incorporation. Distributed control framework for connected automated vehicles  \cite{shi2023physics}. Here information from downstream vehicles and roadway geometry information are incorporated as physics prior knowledge through state fusion.  }
    \label{fig:State_design}
\end{figure}

Vehicular traffic control applications have used physics priors to design the state representations. 
While controlling connected automated vehicles (CAVs), \cite{shi2023physics} proposed the use of surrounding information from downstream vehicles and roadside geometry, by embedding them in the state representation, see Fig.~\ref{fig:State_design}. The physics-informed state fusion approach integrates received information as DRL state (input features) i.e. for the $i^{th}$ CAV, DRL state is given as $s_i^t = \big[ 
e_i^t,\phi_i^t, \delta q_i^{-t}, \delta d_i^{-t}, k_i^{t} \big]$, which are deviation values, (from left): lateral, angular, weighed equilibrium spacing and speed, and road curvature information. 

Jurj et al. \cite{jurj2021increasing} makes use of physical information like jam-avoiding distance to train RL agent, in order to improve collision avoidance of vehicles with adaptive cruise control. In ramp metering control, \cite{han2022physics} utilizes an offline-online policy training process, where the offline training data consists of historical data and synthetic data generated from a physical traffic flow model.

In \cite{cao2023physics} a physics informed graphical representation-enabled, global graph attention (GGAT) network is trained to model power flow calculation process. Informative features are then extracted from the GGAT layer (as representation N/W) and transferred used in the policy training process. While
\cite{gokhale2022physq}, uses PINNs based on thermal dynamics of buildings for learning better heating control strategies. Dealing with 
aircraft conflict resolution problem, \cite{zhao2021physics} composed intruder's information e.g. speed and heading angle into an image state representation. This image now constitutes of the physics prior and serves as the input feature for RL based learning. In \cite{zhang2022barrier}, the authors proposed a safe reinforcement learning algorithm using barrier functions for distributed MPC nonlinear multi-robot systems, with state constraints. \cite{ohnishi2019barrier}, incorporates trained model alongside control barrier certificates, which restrict policies and prohibits exploration of the RL agent into certain undesirable sections of the state space. In case of a safety breach due to non-stationarity, the Lyapunov stability conditions ensures the re-establishment of safety.

\subsubsection{Action regulation:} 

\begin{figure}[h!]
    \centering\includegraphics[width=0.8\linewidth]{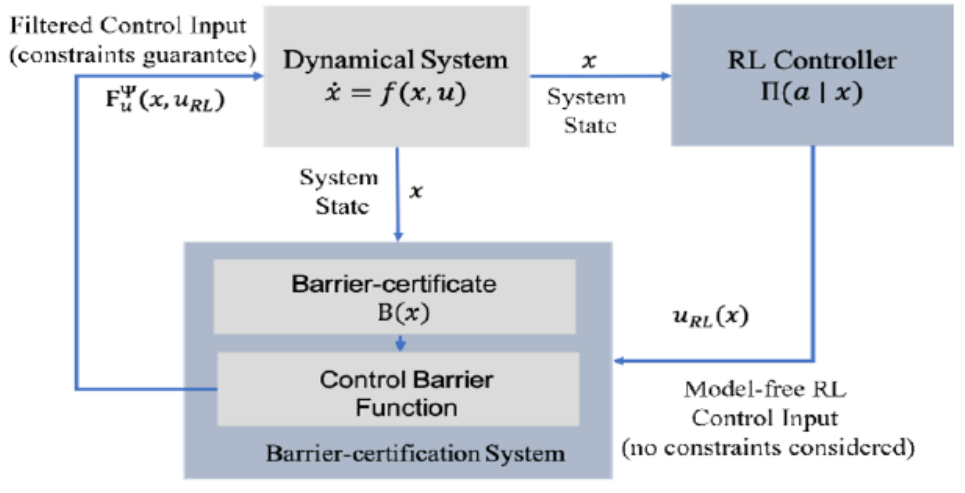}
    \caption{ Example of action regulation, using physics priors. In \cite{zhao2023barrier}, a barrier certification system receives RL control policy generated control actions and refines them sequentially using a barrier certificate to satisfy operational constraints. }
    \label{fig:Action_regulation}
\end{figure}

Many safety critical applications have used physics based constraints and other information in action regulation. These kind of approaches can be categorized under shielded RL/ safety filter, where a type of safety shield or barrier function is employed to check the actions.

For safe power system control \cite{zhao2023barrier} 
proposes a framework for learning a
stabilizing controller that satisfies predefined safety regions, see Fig.~\ref{fig:Action_regulation}. Combining a model-free controller and a barrier-certification system, using a NN based barrier function, i.e. neural barrier certificate (NBC). Given a training set they learn a NBC $B_{\epsilon}(x)$ and filtered (regulated) control action $\mathcal{F}_u^{\psi}$, jointly holding the following condition
\begin{equation*}
    \begin{split}
    (\forall{x} \in \mathcal{S}_{0}, B_{\epsilon}(x) \leq 0 ) \wedge  (\forall{x} \in \mathcal{S}_{u}, B_{\epsilon}(x) > 0 ) \\
    \wedge (\forall{x}\in {x|B_{\epsilon}(x)=0}, \mathcal{L}_{f(x,u_{RL})}B_{\epsilon }(x)<0)
    \end{split}
\end{equation*}

where $\mathcal{L}_{f(x,u_{RL})}B_{\epsilon }(x)$ is the Lie derivative of $B_{\epsilon }(x)$, and $\phi, \epsilon$ are NN parameters. $\mathcal{S}_0, \mathcal{S}_u$ are set of initial states and unsafe states respectively.

\cite{cheng2019end} introduces a hybrid approach of MFRL and MBRL using  CBF, with provision of online learning of unknown system dynamics. It assumes availability of a set of safe states.
In a MARL setting, \cite{cai2021safe} introduced cooperative and non-cooperative CBFs in a collision-avoid problem, which includes both cooperative agents and obstacles. Also in MARL setting, \cite{chen2022physics} proposed efficient active voltage controller of photovoltaics (PVs) enabled with shielding mechanism. Which ensures safe actions of battery energy storage
systems (BESSs) during training. 
\cite{yu2023district}  deals with controlling a district cooling system (DCS), with complex thermal dynamic model and uncertainties from regulation signals and cooling demands. The proposed safe controller a hybrid of barrier function and DRL and helps avoid unsafe explorations and improves training efficiency.
\cite{li2021safe} proposed a safe RL framework for adaptive cruise control, based on a safety-supervision module. 
The authors used the underlying system dynamics and exclusion-zone requirement to construct a safety set, for constraining the learning exploration.

In a highway motion planning setting for autonomous vehicles\cite{wang2022ensuring} proposed a CBF-DRL hybrid approach. 
Certain works like \cite{cao2023physical_1} and
\cite{cao2023physical_2} have introduced multiple physics based artifacts to ensure safe learning in autonomous agents. Both of them used residual control based architecture merging physical model and data driven control. Additionally it also leverages physics model guided reward. \cite{cao2023physical_2} extends the work by \cite{cao2023physical_1} and introduces physics model guided policy and value network editing in addition to the physics based reward.
In \cite{duan2021learning}, the authors integrate learning a task space policy with a model based inverse dynamics controller, which translates task space actions into joint-level controls. This enables the RL policy to learn actions in task space.

\subsubsection{Reward design:}

\begin{figure}[h!]
    \centering\includegraphics[width=0.7\linewidth]{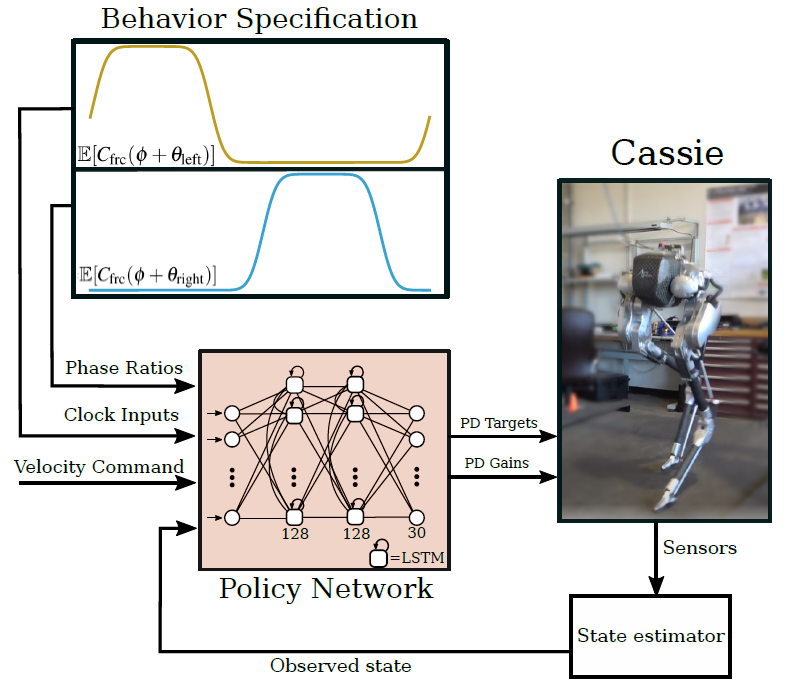}
    \caption{ Example of physics incorporation in reward design. In \cite{siekmann2021sim} a reward function design framework was introduced, that describe robot gaits as a periodic phase sequence such that each of which rewards or penalizes a particular physical system measurement.  }
    \label{fig:Reward_design}
\end{figure}

In sim-to-real setting \cite{siekmann2021sim} proposed a reward specification framework based on composing probabilistic periodic costs on basic forces and velocities, see Fig.~\ref{fig:Reward_design}. The framework defines a parametric reward function for common robotic (bipedal) gaits. Dealing with periodic robot behavior, the absolute time reward function is here defined  in terms of a cycle time variable $\phi$ (which cycles over time period of $[0,1]$, as $R(s,\phi)$. The updated reward function as given below, is defined as a biased sum of $n$ reward components $R_i(s,\phi)$, each capturing a desired robot gait characteristic.
\begin{equation*}
    \begin{split}
        R(s,\phi) = \beta + \Sigma R_i(s,\phi),\; \text{where}\\
        R_i(s,\phi) = c_i \times I_i(\phi)\times q_i(s)
    \end{split}
\end{equation*}
each $R_i(s,\phi)$ is a product of phase-coefficient $c_i$, phase indicator $I_i(\phi)$ and phase reward measurement $q_i(s)$.

In \cite{chen2023hybrid}, the authors introduced a RL-PIDL hybrid framework, to learn MFGs, which generalize well and manage can complex multi-agent systems applications. The physics based reward component (= evolution of population density/ mean-field state) is approximated using PINN.
To better mimic natural human locomotion \cite{korivand2023inertia}, designed reward function based on physical and experimental information: trajectory optimization rewards, and bio-inspired rewards. In a similar task of imitation of human motion but from motion clip, \cite{chentanez2018physics} proposes a physics-based controller using DRL. A rigid body physics simulator is used to solve rigid body poses that closely follows the motion capture (mocap) clip frames.
In a similar work \cite{peng2018deepmimic}, a data driven RL framework was introduced for training control policies for simulated characters. Refernce motions are used to define imitation reward and the task goal defines task specific reward.

\cite{li2023federated} leverages a federated MADRL approach for energy management in multi-microgrid settings. 
The reward is designed to satisfy two physical targets: operation cost and self energy sufficiency. \cite{yousif2023physics} proposed a DRL based method for reconstruction of flow fields from noisy data. Physical constraints like momentum equation, pressure Poisson equation and boundary conditions are used for designing the reward function. \cite{yin2023generalizable} proposed physics based reward shaping for wireless navigation applications. They used a cost function augmented with physically motivated costs like costs for link-state monotonicity, for angle of arrival direction following, and for SNR increasing.
In single molecule 3D structure optimization problem, \cite{cho2019physics} used physics based DFT calculation is used as reward function, for physically correct structural prediction.
In \cite{li2019temporal}, the authors used temporal logic through a finite state automata (FSA), control Lyapunov and barrier function for ensuring effective and safe RL in complex environments.The FSA simultaneously provides rewards, objectives and safety constraints to the framework components.

Addressing the problem of  dexterous manipulation of objects in virtual environments, \cite{garcia2020physics} trained the agent in a residual setting using hybrid model-free RL-IL approach. Using a physics simulator and a pose estimation reward the agent learns to refine the user input to achieve a task while keeping the motion close to the input and the expert demonstrations. \cite{luo2020kinematics} tackles physically valid 3D pose estimation from egocentric video. The authors utilized a combination of kinematics and dynamics approach, whereby the residual of the action against a learned kinematics model is outputted by the dynamics-based model.
In \cite{jiang2022physics},the authors proposed inclusion of physics based intrinsic reward for improved policy optimization of RL algorithms. 

In the context of predicting interfacial area  in two-phase flow, \cite{dang2022towards} proposed. The two-phase flow physics information is infused into the underlying MDP framework, which is then uses RL strategies to describe behavior of flow dynamics. The work introduces multiple rewards based on physical interfacial area transport models, other physical parameters and data. In a work concerning optimization of nuclear fuel assembly \cite{radaideh2021physics}, the authors introduce a reward shaping approach in RL optimization, which is based on physical tactics used by fuel designers.  These tactics include moving fuel rods in assembly to meet certain constraints and objectives.

A number of works have used physics through multiple PIRL methods. Apart from reward design they have infused physics information through state design \cite{shi2023physics} and action regulation \cite{cao2023physical_2,cao2023physical_1, wang2022ensuring}. They have been discussed in previous sections and hence not repeated.

\subsubsection{Augment simulator or model:}

\begin{figure}[htb]
  \centering\includegraphics[width=0.8\linewidth]{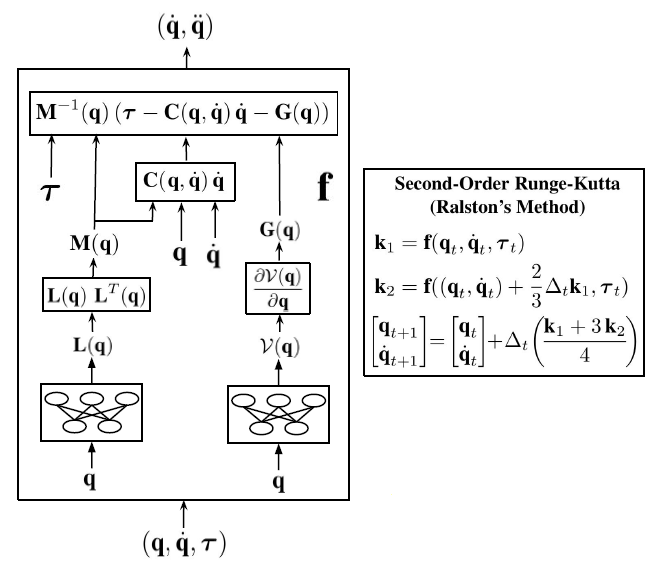}
    \caption{Example, augmentation of learnable model using physics information. The figure shows system dynamics learning network structured using a LNN \cite{ramesh2023physics} and next state calculations using Ralston's method. Here the PINN (LNN) based dynamics model and reward model , are learned via data-driven method. }
    \label{fig:augment_model}
\end{figure}

In MBRL setting, using structure of underlying physics, and building upon Lagrangian neural network (LLN) \cite{cranmer2020lagrangian}, Ramesh et al.\cite{ramesh2023physics} learned the system model via data-driven approach, see Fig.~\ref{fig:augment_model}. Concerning systems obeying Lagrangian mechanics, the state consists of generalized coordinates $q$ and velocities $\dot{q}$. Lagrangian, which is a scalar is defined as
\begin{equation*}
    \begin{split}
        \mathcal{L}(q,\dot{q},t)= \mathcal{T}(q,\dot{q})-\mathcal{V}(q)
    \end{split}
\end{equation*}
where $\mathcal{T}(q,\dot{q})$ is kinetic energy and $\mathcal{V}(q)$ is the potential energy. And so the Lagrangian equation of motion can be written as
\begin{equation*}
    \begin{split}
    \tau &= M(q)\ddot{q}+C(q,\dot{q})\dot{q}+G(q), \text{where}\\
    \ddot{q} &= M^{-1}(q)(\tau - C(q,\dot{q})\dot{q}-G(q))
    \end{split}
\end{equation*}
where $C(q,\dot{q})\dot{q}$ is Coriolis term, $G(q)$ is gravitational term and $\tau$ is motor torque. 
In the NN implementation, separate networks are used for learning $\mathcal{V}(q)$ and $L(q)$, leveraging which the acceleration $(\ddot{q})$ quantity is generated. The output state derivative $(\dot{q},\ddot{q})$ is then integrated using $2^{nd}$-order Runge-Kutta to compute next state.

Concerning a sim-to-real setting, in \cite{golemo2018sim} authors train a recurrent neural network on the differences between robotic trajectories in simulated and actual environments. This model is further used to improve the simulator.
For improved transfer to real environment, \cite{lowrey2018reinforcement} collected hardware data (positions and calculated system velocities) to seed the simulator, for training control policies. 
\cite{alam2021physics} proposes a framework for autonomous manufacturing of acoustic meta-material, while leveraging physics informed RL and transfer learning. A physics guided simulation engine is used to train the agent in source task and then fine-tuned in a data-driven fashion in the target task.

\cite{martin2022reinforcement} introduced a PINN based gravity model for training of dynamically informed RL agents. 
\cite{rodwell2023physics} uses surrogate models that capture primary physics of the system, as a starting point of training DRL agent. In a curriculum learning setting, they train an agent to first track limit cycles in a velocity space for a representative non-holonomic system and then further trained on a small simulation dataset.
\cite{xie2016model} combines linear dynamic models of physical systems with optimism driven exploration. Here the features for the linear models obtained from robot morphology and the exploration is done using MPC.

A number of works introduced novel models are better representations of real world physics and serves as better simulators and ensures effective sim to real transfers. \cite{sanchez2018graph} introduced learnable physics models which supports accurate predictions and efficient generalization across distinct physical systems. 
Concerning dynamic control with partially known underlying physics (governing laws), \cite{liu2021physics} proposed a physics informed learning architecture, for environment model. ODEs and PDEs serves as the primary source of physics for these models.
\cite{veerapaneni2020entity} uses entity abstraction to integrate graphical models, symbolic computation and NNs in a MBRL agent. The framework presents  object-centric perception, prediction and planning which helps agents to generalize to physical tasks not encountered before.
\cite{lee2020context} proposes a context aware dynamics model which is adaptable to change in dynamics. They break the problem of learning the environment dynamics model into two stages: learning context latent vector and predicting next state conditioned on it.

In micro-grid power control problem, \cite{she2023inverter} combines model-based analytical proof and reinforcement learning. Here model-based derivations are used to narrow the learning space of the RL agen, reducing training complexity significantly.
In visual model based RL, \cite{veerapaneni2020entity} models a scene in terms of entities and their local interactions, thus better generalizing to physical task the learner has not seen before. Similar to learning entity abstractions, in \cite{lee2020context} the authors tackles the challenge of learning a generalizable global model through: learning context latent vector, capturing local dynamics and predicting next state conditioned on the encoded vector.
Addressing dynamic control problem in MBRL setting, \cite{liu2021physics} leveraged physical laws (in form of canonical ODE/ PDE) and environmental constraints to mitigate model bias issue and sample inefficiency.
In autonomous driving safe ramp merging problem, \cite{udatha2022safe} embedded probabilistic CBF in RL policy in order to learn safe policies, that also optimize the performance of the vehicle. Typically CBFs need good approximation of car's model. Here the probabilistic CBF is used as an estimate of the model uncertainty.

\cite{cho2019physics} incorporates physics through reward design as well as through simulator augmentation, and has been discussed in previous section.

\subsubsection{Augment policy and/or value N/W:}

\begin{figure}[htb]
  \centering\includegraphics[width=0.9\linewidth]{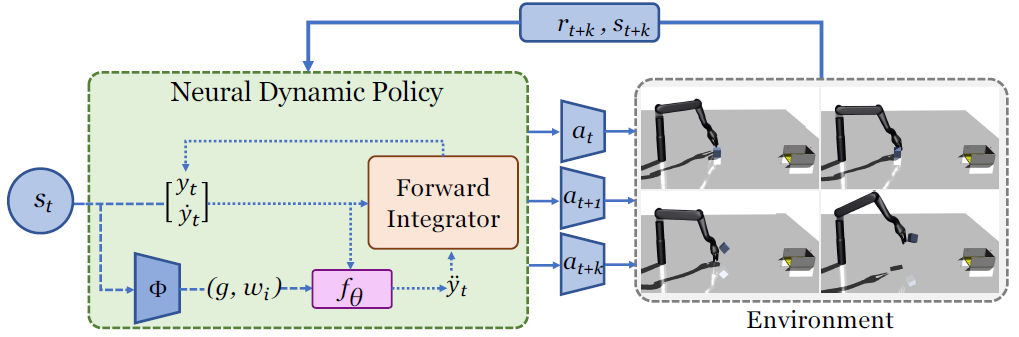}
    \caption{Example, augmentation of policy using physics information. In \cite{bahl2020neural}, given an observation $s_t$ from the environment, a neural dynamic policy generates $w$ i.e. the weights of basis function and $g$ which is a goal for the robot, for a function $f_{\theta}$. This function is then used by an open loop controller to generate a set of actions from the robot to execute in the environment and collect next states and rewards to train the policy. }
    \label{fig:augment_policy}
\end{figure}

In \cite{bahl2020neural}, Bahl et al. proposes Neural Dynamic Policies (NDP) where they incorporate dynamical system as a differentiable layer in the policy network, see Fig.~\ref{fig:augment_policy}. In NDP, a NN $\Phi$ takes an input state $(s_t)$ and predicts parameters of the dynamical system (i.e. $(w,g)$ ). Which are then used to solve second-order differential equation $\ddot{y} = \alpha(\beta(g-y)-\dot(y))+f(x)$, to obtain system states $(y,\dot{y},\ddot{y}).$, which represents the behavior of the dynamic system, given a state goal $g$. Here $\alpha, \beta$ are global parameters allowing critical damping of system and $f$ is a non-linear forcing function which primarily captures the shape of trajectory.
Depending on robot's coordinate system an inverse controller may also be used to convert $y$ to $a$, i.e. $a = \Omega(y,\dot{y},\ddot{y})$
. The NDPs thus can be defined as 
\begin{equation*}
    \begin{split}
        \pi(a|s;\theta) &\triangleq \Omega(DE(\Phi(s;\theta))),\text{where}\\
        DE(w,g) &\rightarrow \{y,\dot{y},\ddot{y}\}
    \end{split}
\end{equation*}
here $DE(w,g)$ represents solution of the differential equation.

Extending this work to hierarchical deep policy learning framework, \cite{bahl2021hierarchical} introduced H-NDP which forms a curriculum by learning local dynamical system-based policies on small state-space region and then refines them into global dynamical system based policy.
Given the accurate dynamics and constraint of the system \cite{zhao2022safe} introduces control barrier certificates into actor-critic RL framework, for learning safe policies in dynamical systems.
\cite{margolis2021learning} proposes a method for generating highly agile and visually guided locomotion behaviors. They leverage MFRL while using model based optimization of ground reaction forces, as a behavior regularizer. 

In \cite{du2023reinforcement} proposes an approach of safe exploration using CLBF without explicitly employing any dynamic model. The approach approximate the RL critic as a  CLBF, from data samples and parameterized with DNNs. Both the actor and critic satisfies reachability and safety guarantees. 
\cite{mukherjee2023bridging} combines PINN with RL, where the value function is treated as a PINN to solve Hamilton-Jacobi-Bellman (HJB) PDE. It enables the RL algorithm to exploit the physics of environment aswell as optimal control to improve learning and convergence.

\cite{park2023physics} proposes an optimization method for freeform nanophotonic devices, by combining adjoint based methods (ABM) and RL. In this work the value network is initialized with adjoint gradient predicting network during initialization of RL process.
Cao et al. \cite{cao2023physical_2} have used physics model to influence reward function, as well as edit policy and value networks as necessary. The work has been mentioned before in reward design.
%

To improve policy optimization, \cite{mora2021pods} used differentiable simulators to directly compute the analytic gradient of the policy's value function w.r.t. the actions generated by it. This gradient information is used to monotonically improve the policy's value function.
Gao et al. \cite{gao2022transient} proposes a transient voltage control approach, by integrating physical and data-driven models of power system. They also uses the constraint of the physical model on the data-driven model to speed up convergence. A PINN trained using PDE of transient process acts as the physical model and contributes directly to the loss of the RL algorithm.

Xu et al. \cite{xu2022accelerated} presents an efficient differentiable simulator (DS) with a new policy training algorithm which can effectively leverage simulation gradients. The learning algorithm alleviates issues inherent in DS while allowing many physical environments to be run in parallel.
%
\cite{chen2022physics} incorporates physics through action regulation and penalty signal to agent, and has been discussed in previous section. 

In MBRL setting, \cite{lv2022sam} leverage differentiable physics-based simulation and differentiable rendering. By comparing raw observations between simulated and real world, the initial learned system model is continually updated, producing a more physically consistent model.
In data center (DC) cooling control application, \cite{wang2023phyllis} proposed a lifelong-RL approach under evolving DC environment. It leverages physical laws of thermodynamics and the system and models the DC thermal transition and power usage through data collected online. Utilizing learned state transition and reward models it accelerates online adaptation.

Working with a nominal system model, \cite{choi2020reinforcement} presented an RL framework where the agent learns model uncertainty in multiple general dynamic constraints, e.g. CLF and CBF, through data-driven training. A quadratic program then solves for the control that satisfies the safety constraints under learned model uncertainty.

\begin{figure}[t]
\centering

\subf{\includegraphics[width=0.9\columnwidth]{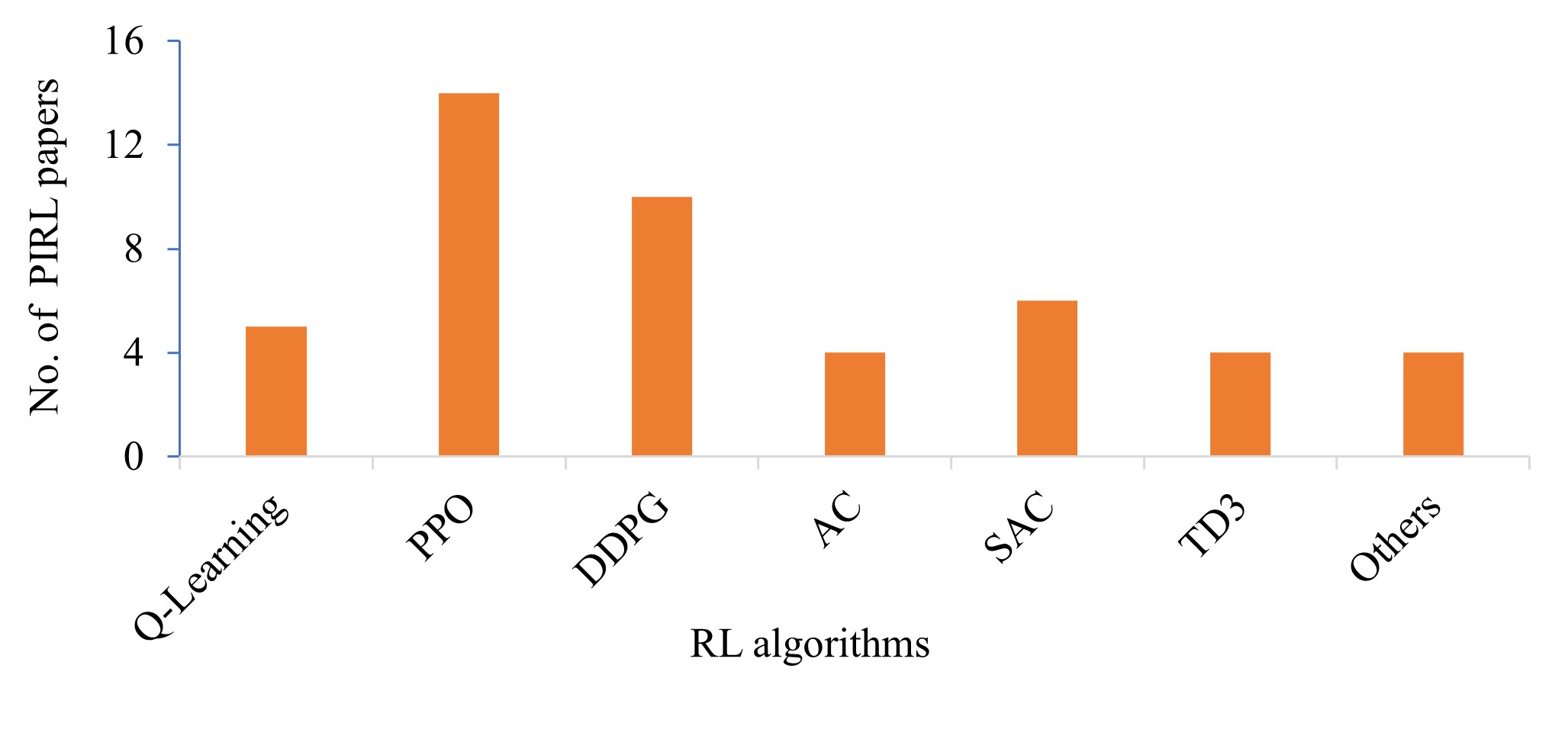}}
     {(a)}
     \label{fig:RL_algo_use}
       
\subf{\includegraphics[width=0.95\columnwidth]{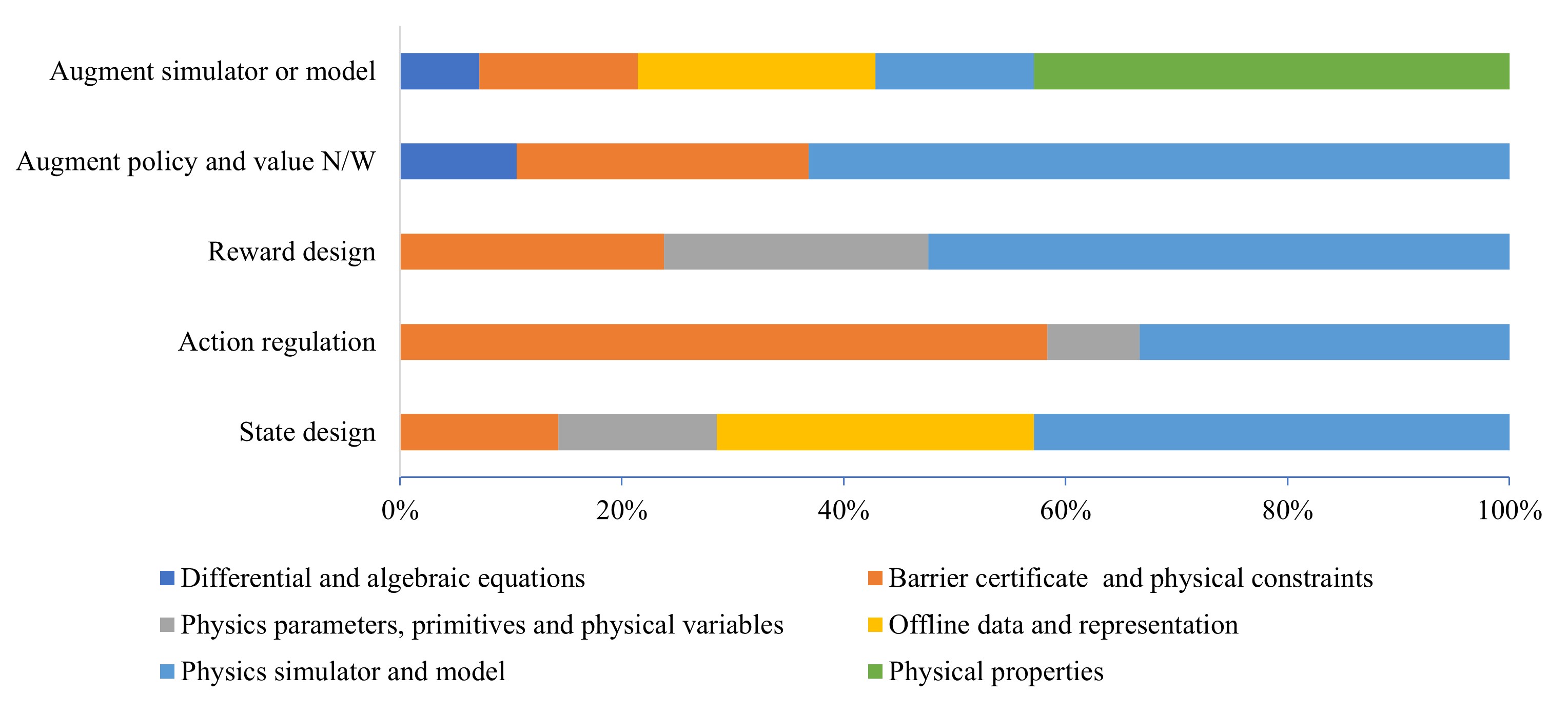}}
     {(b)}
     \label{fig:pi_type_methods}
     
\subf{\includegraphics[width=0.9\columnwidth]{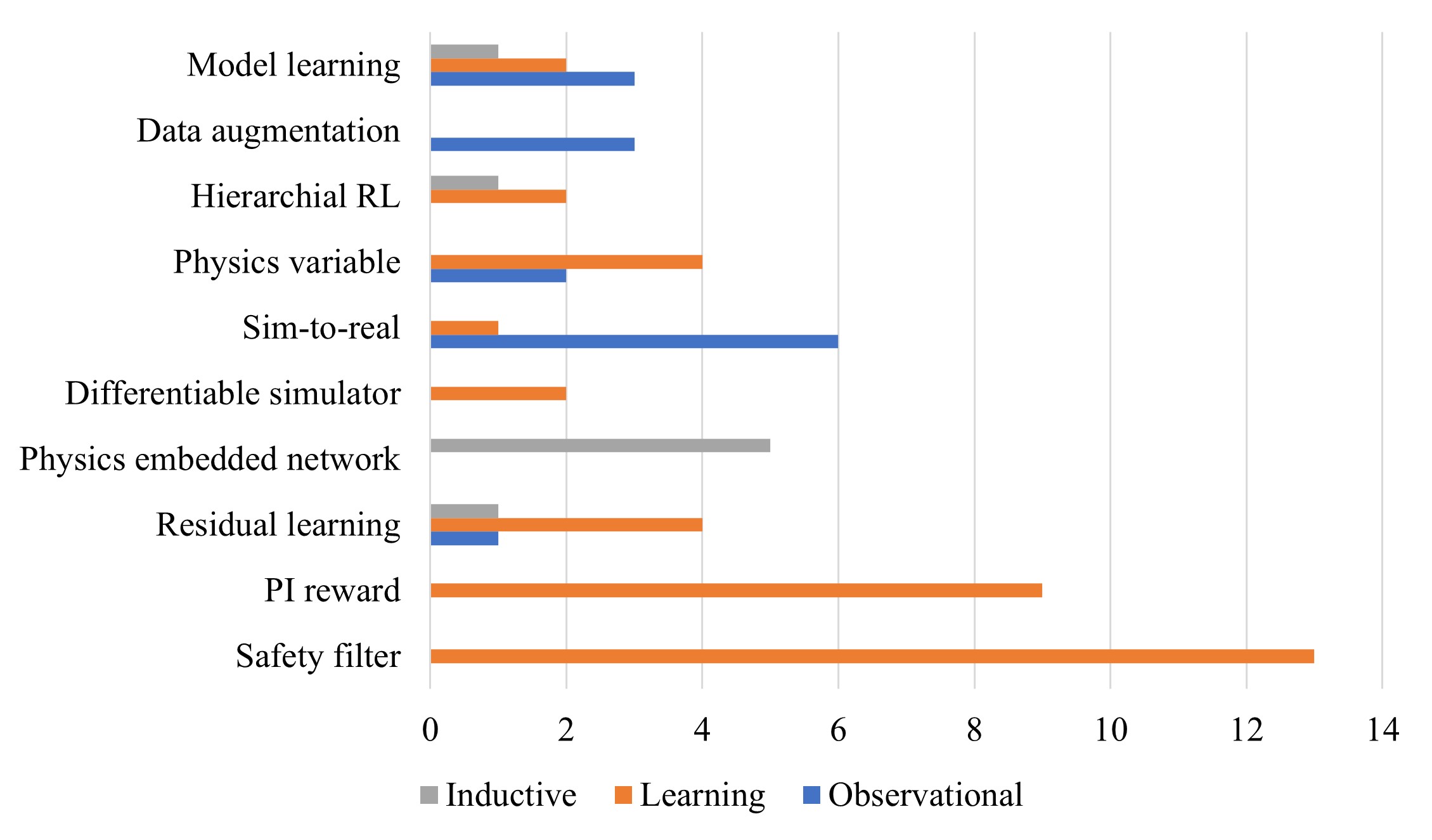}}
     {(c)}
     \label{fig:lrn_arch_bias}

\subf{\includegraphics[width=0.9\columnwidth]{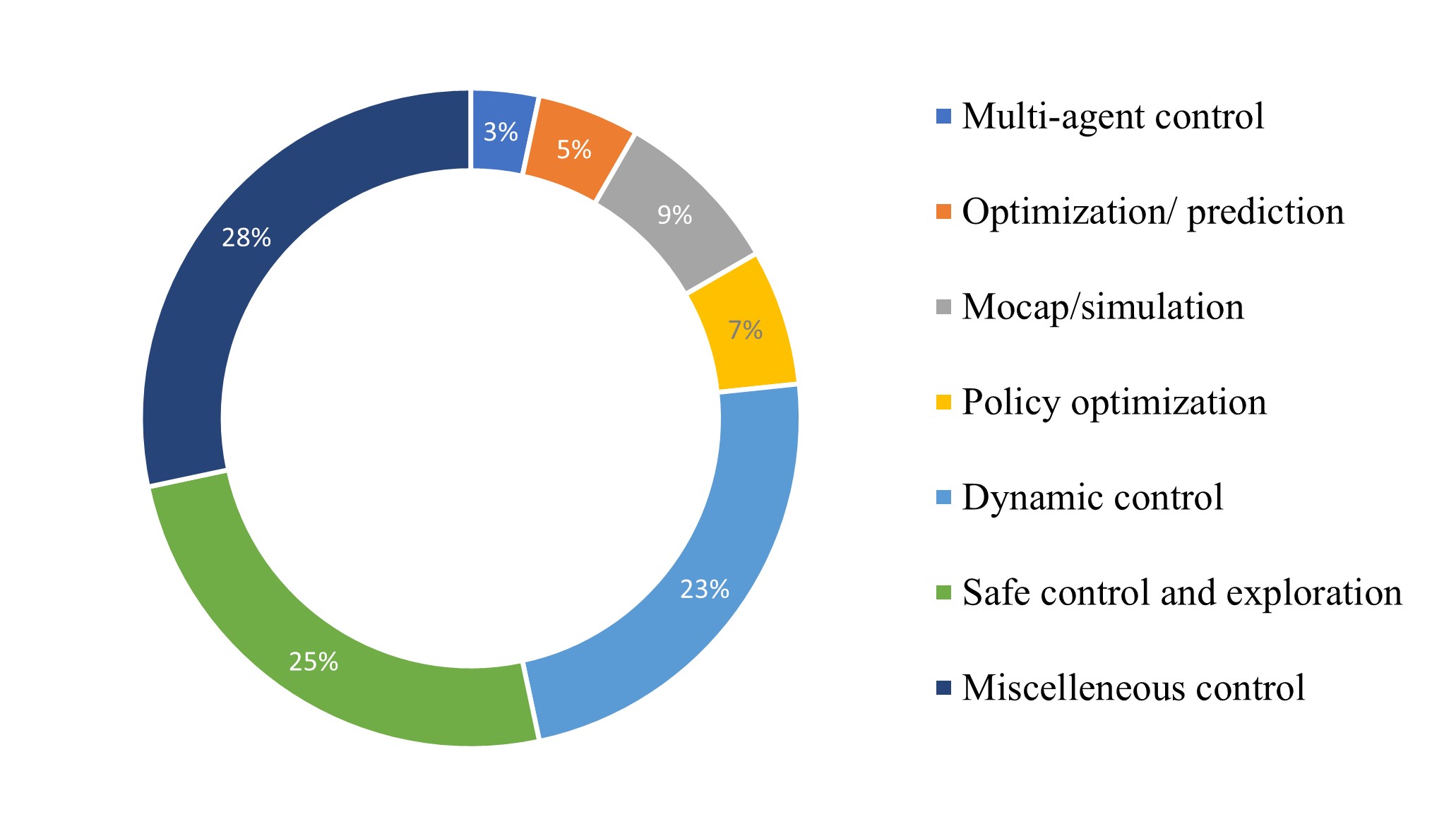}}
     {(d)}
     \label{fig:App_domain}

\caption{Statistical analysis of PIRL literature. (a) Statistic of type of RL algorithms used, (b) Statistic of PI types used in each PIRL method, (c)  Statistic of PIRL learning architectures and related biases, (d)  Statistic of PIRL applications in different domains.} 
\label{fig:analysis}
\end{figure}

\renewcommand{\arraystretch}{1.1}
\begin{table*}[htpb]
\centering
\caption{Summary of PIRL literature - Model Free. } \label{table: PIRL characteristics} 
\resizebox{\textwidth}{!}{%
\begin{tabular}{|l|l|l|l|l|l|l|l|l|l|}
\hline
%
\textbf{Ref.}& \textbf{Year} &\textbf{Context/ Application} & \textbf{RL Algorithm}&\textbf{Learning arch.} & \textbf{Bias} & \textbf{Physics information} & \textbf{PIRL methods} &  \textbf{RL pipeline }\\
\hline
& & & & & & & &\\
\cite{chentanez2018physics}& 2018 & Motion capture & PPO & Physics reward & Learning & Physics simulator & Reward design & Problem representation\\
\cite{peng2018deepmimic} & 2018 & Motion control  & PPO \cite{schulman2017proximal} & Physics reward & Learning & Physics simulator & Reward design & Problem representation\\
\cite{golemo2018sim} & 2018 & Policy optimization & PPO  & Sim-to-Real & Observational & Offline data  & Augment simulator & Training\\
\cite{lowrey2018reinforcement} & 2018 & Policy optimization & NPG \cite{williams1992simple} (C)$^*$ & Sim-to-Real & Observational & Offline data  & Augment simulator & Training\\
%
%
%
& & & & & & & &\\
\cline{1-9}
& & & & & & & &\\
\cite{cho2019physics} & 2019 & Molecular structure optimization & DDPG & Physics reward & Learning & DFT (PS) & Reward design & Problem representation \\
& & & & & & & Augment simulator & Training\\
\cite{li2019temporal} & 2019 & Safe exploration and control & PPO & Residual RL & Learning & CBF, CLF, FSA/TL (BPC) & Reward design &  Problem representation\\
& & & & & & & Augment policy & Learning strategy\\
& & & & & & & &\\
\cline{1-9}
 & & & & & & & &\\
%
%
\cite{bahl2020neural} & 2020 & Dynamic system control & PPO & Phy. embed. N/W & Inductive & DMP (PPV) & Augment policy & Network design \\
\cite{garcia2020physics} & 2020 & Dexterous manipulations & PPO & Residual RL & Observational & Physics simulator & Reward design & Problem representation \\
\cite{luo2020kinematics} & 2020 & 3D Ego pose estimation & PPO & Physics reward & Learning & Physics simulator & State, Reward design & Problem representation \\
%
%
& & & & & & & &\\
\cline{1-9}
 & & & & & & & &\\
\cite{bahl2021hierarchical} & 2021 & Dynamic system control & PPO & Hierarchical RL & Inductive & DMP (PPV) & Augment policy & Network design \\
\cite{margolis2021learning} & 2021 & Dynamic system control & PPO & Hierarchical RL & Learning & WBIC (PPV) & Augment policy & Learning strategy\\
\cite{alam2021physics} & 2021 & Manufacturing & SARSA \cite{sutton1998reinforcement}
& Sim-to-Real & Observational& Physics engine & Augment simulator & Training\\
\cite{siekmann2021sim} & 2021 & Dynamic system control & PPO & Phy. variable & Learning & Physics parameters & Reward design & Problem representation\\

\cite{li2021safe} &  2021 & Safe exploration and control & NFQ \cite{riedmiller2005neural} &  Safety filter & Learning & Physical constraint & Action regulation & Problem representation\\
%
%
\cite{jurj2021increasing} & 2021 & Safe cruise control & SAC & Phy. variable  & Observational & Physical state (PPV) & State design & Problem representation\\
%
%
\cite{mora2021pods} & 2021 & Policy optimization & DPG (C) & Diff. Simulator & Learning & Physics simulator & Augment policy  & Learning strategy\\
\cite{radaideh2021physics} & 2021 & Optimization, nuclear engineering & DQN, PPO & Physics reward & Learning bias & Physical properties (PPR) & Reward design & Problem representation\\
\cite{zhao2021physics} & 2021 & Air-traffic control & PPO  & Data augmentation & Observational & Representation (ODR) & State design & Problem representation\\
 & & & & & & & &\\
\cline{1-9}
& & & & & & & &\\
\cite{wang2022ensuring} & 2022 & Motion planner & PPO + AC \cite{konda1999actor} & Safety filter & Learning & CBF (BPC) & Action regulation & Problem representation\\
& & & & & & & Reward design &\\
\cite{chen2022physics} & 2022 & Active voltage control & TD3 (C)& Safety filter & Learning & Physical constraints  & Penalty function & Problem representation \\
 & & & & & & & Action regulation &  \\
\cite{dang2022towards} & 2022 & Interfacial structure prediction & DDPG & Off-policy & Learning & Physics model & Reward design & Problem representation\\
\cite{gao2022transient} & 2022 & Transient voltage control & DQN & PINN loss & Learning & PDE (DAE) & Augment policy & Learning strategy\\
\cite{gokhale2022physq} & 2022 & Building control & Q-learning (C)  & Data augment & Observational & Representation (ODR)& State design  & Problem representation\\
\cite{han2022physics} & 2022 & Traffic control & Q-Learning & Data augment & Observational & Physics model & State design & Problem representation\\
\cite{martin2022reinforcement} & 2022& Safe exploration and control & SAC & Sim-to-Real & Observational & Physics model & Augment simulator & Training\\
\cite{jiang2022physics} & 2022 & Dynamic system control & SAC (etc.)& Physics reward & Learning & Barrier function & Reward design & Problem representation\\
\cite{xu2022accelerated} & 2022 & Policy Learning & Actor-critic (C)& Diff. Simulator & Learning & Physics simulator & Augment policy & Learning strategy\\
& & & & & & & &\\

\cline{1-9}
& & & & & & & &\\
\cite{cao2023physical_1} & 2023 & Safe exploration and control & DDPG & Residual RL & Learning & Physics model & Reward design & Problem representation\\
& & & & & & & Action regulation  &\\
\cite{cao2023physical_2} & 2023 & Safe exploration and control & DDPG & Residual RL & Inductive & Physics model & Reward design & Problem representation\\
 & & & & & & & Action regulation  & \\
& & & & & Inductive & & N/W editing (Aug. pol.) & Network design\\
\cite{cao2023physics} & 2023 & Robust voltage control & SAC & Data augment & Observational & Representation (ODR) & State design & Problem representation\\
\cite{chen2023hybrid} & 2023  & Mean field games & DDPG & Physics reward & Learning & Physics model & Reward design & Problem representation\\
\cite{yang2023model} & 2023 & Safe exploration and control & PPO (C) & Safety filter & Learning & NBC (BPC) & Augment policy & Training \\
\cite{zhao2023barrier} & 2023 & Power system stability enhancement & Custom & Safety filter & Learning & NBC (BPC) & Action regulation & Problem representation \\
\cite{du2023reinforcement} & 2023 &  Safe exploration and control & AC (C) & Safety filter  & Learning  & CLBF \cite{romdlony2016stabilization, dawson2022safe} (BPC) & Augment value N/W & Training \\
\cite{shi2023physics} & 2023 & Connected automated vehicles & DPPO & Physics variable & Observational & Physical state (PPV) & State design & Problem representation\\
&& & & & Learning & &  Reward design  &\\
\cite{korivand2023inertia} & 2023 & Musculoskeletal simulation & SAC (C) & Physics variable & Learning & Physical value & Reward design & Problem representation\\
\cite{li2023federated} & 2023 & Energy management & MADRL(C) & Physics variable & Learning & Physical target & Reward design & Problem representation\\
%
%
\cite{mukherjee2023bridging} & 2023 &  Policy optimization  & PPO & Phy. embed N/W & Inductive & PDE (DAE) & Augment value N/W & Network design\\
\cite{yousif2023physics} & 2023 & Flow field reconstruction & A3C & Physics reward & Learning & Physical constraints & Reward design & Problem representation\\
\cite{park2023physics} & 2023 &  Freeform nanophotonic devices & $\epsilon -$ greedy Q & Phy. embed N/W & Inductive & ABM & Augment value N/W & Network design\\
\cite{rodwell2023physics} & 2023  & Dynamic system control & DPG & Curriculum learning & Learning & Physics model & Augment simulator & Training\\
\cite{she2023inverter} & 2023 & Energy management & TD3 & Sim-to-Real & Observational  & Physics model & Augment simulator & Learning strategy\\
%
%
%
\cite{yin2023generalizable} & 2023 & Robot wireless navigation & PPO & Physics reward & Learning &  Physical value & Reward design & Problem representation\\
%
%
%
& & & & & & & &\\

\hline
\hline
\end{tabular}}
\label{table:PIRL_Characteristic_MF}
\begin{flushleft}
\vspace{-0.5mm}
{ C$^*$ represents custom versions of the adjacent conventional algorithms.
}
\end{flushleft}
\end{table*}
%
%

%
%
\renewcommand{\arraystretch}{1.1}
\begin{table*}[htpb]
\centering
\caption{Summary of PIRL literature - Model based} \label{table: PIRL characteristics_MB} 
\resizebox{\textwidth}{!}{%
\begin{tabular}{|l|l|l|l|l|l|l|l|l|l|}
\hline
%
\textbf{Ref.}& \textbf{Year} &\textbf{Context/ Application} & \textbf{ Algorithm}&\textbf{Learning arch.} & \textbf{Bias} & \textbf{Physics information}& \textbf{PIRL method} &  \textbf{RL pipeline }\\
\hline
& & & & & & & &\\
\cite{xie2016model} & 2016 & Exploration and control &  - & Model learning & Observational & Sys. morphology (PPR) & Augment model &  Learning strategy\\
& & & & & & & &\\
\cline{1-9}
& & & & & & & &\\
%
%
\cite{sanchez2018graph} & 2018 & Dynamic system control & - & Model learning & Inductive & Physics model & Augment model & Learning strategy\\ 
\cite{ohnishi2019barrier} & 2019 & Safe navigation & - & Safety filter & Learning & CBC (BPC) & Action regulation & Problem representation\\
\cite{cheng2019end} & 2019 & Safe exploration and control & TRPO, DDPG & Residual RL & Learning & CBF (BPC) & Action regulation & Problem representation\\
& & & & & & & &\\
\cline{1-9}
& & & & & & & &\\
\cite{veerapaneni2020entity} & 2020 & Control (visual RL) & - & Model learning & Observational & Entity abstraction (ODR) &  Augment model & Learning strategy\\
\cite{lee2020context} & 2020 & Dynamic system control & - & Model learning & Observational & Context encoding (ODR) & Augment model & Learning strategy\\
\cite{choi2020reinforcement} & 2020 & Safe exploration and control & DDPG \cite{silver2014deterministic} & Safety filter & Learning & CBF, CLF, QP (BPC) & augment policy & Learning strategy \\
%
%
& & & & & & & &\\
\cline{1-9}
& & & & & & & &\\
\cite{liu2021physics} & 2021 & Dynamic system control & Dyna + TD3(C)$^*$ & Model identification & Learning & PDE/ ODE, BC (DAE) & Augment model & Learning strategy\\
%
%
%
\cite{duan2021learning} & 2021 & Dynamic system control & PPO &  Residual-RL & Learning & Physics model & Action regulation & Problem representation\\
%
%
\cite{cai2021safe} & 2021 & Multi agent collision avoidance & MADDPG (C) & Safety filter  & Learning & CBF (BPC) & Action regulation & Problem representation\\
& & & & & & & &\\
\cline{1-9}
& & & & & & & &\\
\cite{lv2022sam} & 2022  & Dynamic system control & TD3(C) & Sim-to-Real & Learning & Physics simulator & Augment policy & Learning strategy\\
\cite{udatha2022safe} & 2022 & Traffic control & AC\cite{ma2021model} & Safety filter  & Learning & CBF (BPC) & Augment model & Learning strategy\\
%
%
\cite{zhao2022safe} & 2022 & Safe exploration and control & DDPG & Safety filter & Learning & CBC (BPC) & Augment policy & Learning strategy\\
\cite{zhang2022barrier} & 2022 & Distributed MPC & AC \cite{jiang2020cooperative} & Safety filter & Learning &  CBF (BPC) & State design & Problem representation\\
& & & & & & & &\\
\cline{1-9}
& & & & & & & &\\
\cite{ramesh2023physics} & 2023 & Dynamic system control & Dreamer \cite{hafner2019dream} & Phy. embed. N/W & Inductive & Physics model & Augment model & Network design\\
\cite{cohen2023safe} & 2023 & Safe exploration and control & - & Safety filter  & Learning & CBF (BPC) & Augment model & Learning strategy\\
\cite{huang2023symmetry} & 2023 & Attitude control & - & Phy. embed N/W & Inductive & System symmetry (PPR)  & Augment model & Network design\\
\cite{wang2023phyllis} & 2023 & Data center cooling & SAC & Model identification & Learning & Physics laws (PPR) & Augment model & Learning strategy\\
%
%
\cite{yu2023district} & 2023 & Cooling system control & DDPG & Residual RL & Learning & CBF (BPC) & Action regulation & Problem representation\\
& & & & & & & & \\
\hline
\hline
\end{tabular}}
\label{table:PIRL_Characteristics_MB}
\begin{flushleft}
\vspace{-0.5mm}
{ C$^*$ represents custom versions of the adjacent conventional algorithms.
}
\end{flushleft}
\end{table*}

\subsection{Review of simulation/ evaluation benchmarks}
In Table~\ref{table:PIRL_evaluation_Benchmarks}, we present the different training and evaluation benchmarks that has been used in the reviewed PIRL literature. We list the important insights from the table:
\begin{itemize}
    \item [1.] A majority works dealing with dynamic control have used OpenAI Gym \cite{xie2016model}, Safe Gym \cite{yang2023model}, MuJoCo \cite{veerapaneni2020entity,zhou2018environment}, Pybullet \cite{du2023reinforcement} and Deep mind control suite environments \cite{sanchez2018graph,ramesh2023physics}, which are standard benchmarks in RL . Works dealing specifically with traffic management have used platforms like SUMO \cite{wang2022ensuring} and CARLA \cite{udatha2022safe}.
    \item [2.] Works dealing with power and voltage management problems have used IEEE distribution system benchmarks \cite{chen2022physics,gao2022transient} to evaluate proposed algorithms. Alternatively in some works MATLAB/ SIMULINK platform is also used for training or evaluating RL agents \cite{she2023inverter}
    \item [3.] One crucial observation is that a huge number of work have used customized or adapted environments for training and evaluation and have not used conventional  environments \cite{li2019temporal,cohen2023safe,lutter2021differentiable}. 
\end{itemize}

\subsection{Analysis}
\subsubsection{Research trend and statistics}
\textbf{Use of RL algorithms: } As is evident from Fig.\ref{fig:analysis} (a), PPO\cite{schulman2017proximal} and its variants are the most preferred RL algorithm, followed by DDPG \cite{silver2014deterministic}. Among the comparatively new algorithms SAC\cite{haarnoja2018soft} is preferred over TD3\cite{fujimoto2018addressing}. 

\textbf{Types of physics priors used: }
In Fig.\ref{fig:analysis} (b), we can see that physics information takes the form of physics simulator, system models, barrier certificates and physical constraints, in a majority of works. PI types \say{Barrier certificate constraints and physical constraint} and \say{Physics simulator and models} dominates in more that $60\%$ of works in \say{Action regulation} and \say{Augment policy and value N/W} PIRL methods.

\textbf{ Learning architecture and bias: } 
In Fig.\ref{fig:analysis} (c) we visualize the relationship between PIRL learning architectures (sec: \ref{Sec: Learning architecture }) and the three biases through which physics is typically incorporated in PIML approaches. In architectures \say{PI reward} and \say{safety filter}, physics is incorporated strictly through \say{learning bias}, signifying the heavy use of constraints, regularizers and specialized loss functions. While \say{Physics embedded network} incorporates physics information through \say{inductive bias}, i.e. through imposition of hard constraints through use specialized and custom physics embodied networks.

\textbf{Application domains: } In Fig.\ref{fig:analysis} (d) almost $85\%$ of the application problem dealt with PIRL approaches relates to controller or policy design. \say{Miscellaneous control} includes optimal policy/ controller learning approaches for different application sectors like  energy management \cite{li2023federated,she2023inverter} and  data-center cooling \cite{wang2023phyllis}, and accounts to majority of applications. 
\say{Safe control and exploration}, includes those works concerning with safety critical systems, ensuring safe exploration and policy learning, accounts for $25\%$. 
\say{Dynamic control}, includes control of dynamic systems, including robot systems and amounts to about $23\%$ of all works surveyed. Other specific applications include optimization/ prediction \cite{cho2019physics,dang2022towards}, motion capture/simulation \cite{wang2022ensuring,chentanez2018physics} and improvement of general policy optimization approaches \cite{golemo2018sim,lowrey2018reinforcement} through physics incorporation.

\subsubsection{RL challenges addressed}\label{Sec: RL_challenges}
In this section we will discuss and elaborate on how recent physics incorporation in RL algorithms have addressed certain open problems of the RL paradigm.
\begin{itemize}
    \item [1)] \textit{Sample efficiency:} RL approaches need a huge number of agent-environment interaction and related data to work.
    One effective way of dealing with this problem is to use a surrogate for the real environment in the form of a simulator or learned model via data-driven approaches. 
    
    PIRL approaches incorporate physics to augment simulators thus reducing the sim-to-real gap, thereby bringing down online evaluation cycles \cite{lv2022sam,alam2021physics}. Also physics incorporation during system identification or model learning phase in MBRL help reduce sample efficiency through learning a truer to real environment using lesser training samples \cite{sanchez2018graph,veerapaneni2020entity}.
    \item [2)] \textit{Curse of dimensionality:} RL algorithms become less efficient both in training and performing on environment defined with high-dimensional and continuous state and action spaces, known as the 'curse of dimensionality'. Typically dimensionality reduction techniques are used to encode the large state or action vectors into low dimensional representations. The RL algorithm is then trained in this low dimensional setting.
    
    PIRL approaches extract underlying physics information from environment through learning physically relevant low dimensional representation from high dimensional observation or state space \cite{gokhale2022physq,cao2023physics}.
    In \cite{gokhale2022physq}, a PINN is utilized to extract physically relevant information about the system's hidden state, which is then used to learn a Q-function for policy optimization.

    \item [3)] \textit{Safety exploration:} Safe reinforcement learning involves learning control policies that guarantee system performance and respect safety constraints during both exploration and policy deployment.

    In safety-critical applications using reinforcement learning, it's crucial to regulate agent exploration. Control Lyapunov function (CLF)\cite{li2019temporal, choi2020reinforcement}, barrier certificate/ barrier function (BF), control barrier function/ certificate (CBF/ CBC) \cite{cheng2019end,cai2021safe} are commonly used concepts. Barrier certificates define safe states, while control barrier functions ensure states stay in the safety set. These approaches are typically used for systems with partial or learnable dynamics model and generally a known set of safe states/ actions.

    \item [4)] \textit{Partial observability or imperfect measurement:} Partial observability is a setting where due to noise, missing information, or outside interference, an RL agent is unable to obtain the complete states needed to understand the environment.

    PIRL approaches modify or enhance the state representation to provide more useful information, in cases of missing or inadequate information. This may involve state fusion, which incorporates additional physics or geographical information from the environment  \cite{jurj2021increasing} or other agents \cite{shi2023physics}.
    
    \item [5)] \textit{Under-defined reward function:} Defining the reward function is critical in creating MDPs and ensuring the effectiveness and efficiency of RL algorithms. However, since they are created by humans, there is a risk of them being under-defined and not guiding the RL algorithm effectively in policy optimization.

    PIRL approaches introduce physics information through effective reward design or augmentation of existing reward functions with bonuses or penalties \cite{dang2022towards,luo2020kinematics,garcia2020physics,siekmann2021sim}.
    For example, in a sim-to-real setting, \cite{siekmann2021sim} proposed a framework for specifying rewards that combines probabilistic costs associated with primary forces and velocities. The framework creates a parametric reward function for common robotic gaits, in biped robots.
\end{itemize}

\renewcommand{\arraystretch}{1.1}
\begin{table}[htpb]
\centering
\caption{Summary of PIRL training/ evaluation benchmarks.}
\resizebox{\linewidth}{!}{%
\begin{tabular}{|l|l|l|}
\hline
\textbf{Simulator/ platform} & \textbf{Specific environment/ system name} &
\textbf{Reference} \\
\hline
OpenAI Gym & Pusher, Striker, ErgoReacher & \cite{golemo2018sim} \\
OpenAI Gym & Mountain Car, Lunar Lander $(continuous)$ & \cite{jiang2022physics}\\
OpenAI Gym & Cart-Pole, Pendulum (simple and double)  &\cite{xie2016model} \\
OpenAI Gym & Cart-pole & \cite{cao2023physical_1} \\
OpenAI Gym & Cart-pole and Quadruped robot & \cite{cao2023physical_2} \\
OpenAI Gym  & CartPole, Pendulum & \cite{liu2021physics}\\
OpenAI Gym & Inverted Pendulum $(pendulum-v0)$, & \cite{cheng2019end}\\
OpenAI Gym & Mountain car $(cont.)$, Pendulum, Cart pole & \cite{zhao2022safe}\\
OpenAI Gym & Simulated car following \cite{he2018data}&\\
MuJoCo & Ant, HalfCheetah, Humanoid, Walker2d & \cite{mukherjee2023bridging} \\
& Humanoid standup, Swimmer, Hopper &\\
& Inverted and Inverted Double Pendulum $(v4)$ &\\
MuJoCo & Cassie-MuJoCo-sim \cite{cassie_mujoco} & \cite{siekmann2021sim,duan2021learning} \\
& 6 DoF Kinova Jaco \cite{ghosh2017divide} & \cite{bahl2020neural,bahl2021hierarchical} \\
MuJoCo & HalfCheetah,
Ant,& \cite{lee2020context}\\
& CrippledHalfCheetah, and SlimHumanoid \cite{zhou2018environment}&\\
MuJoCo & Block stacking task \cite{janner2018reasoning} & \cite{veerapaneni2020entity}\\
OpenAI Gym & CartPole, Pendulum &\\
OpenSim-RL\cite{kidzinski2018learning} & L2M2019 environment &  \cite{korivand2023inertia}\\
Safety gym \cite{yuan2021safe} & Point, car and Doggo goal& \cite{yang2023model}\\
- & Cart pole swing up, Ant & \cite{xu2022accelerated}\\
- & Humanoid, Humanoid MTU &\\
- & Autonomous driving system & \cite{chen2023hybrid}\\
Deep control suite \cite{tassa2018deepmind} & Pendulum, Cartpole, 
Walker2d & \cite{sanchez2018graph}\\
& Acrobot, Swimmer, Cheetah & \\
- & JACO arm (real world) &\\
Deep control suite & Reacher, Pendulum, Cartpole, & \cite{ramesh2023physics} \\
& Cart-2-pole, Acrobot, &\\
& Cart-3-pole and Acro-3-bot &\\
-& Rabbit\cite{chevallereau2003rabbit} & \cite{choi2020reinforcement}\\
MARL env. \cite{lowe2017multi} & Multi-agent particle env.& \cite{cai2021safe}\\
ADROIT\cite{rajeswaran2017learning} & Shadow dexterous hand & \cite{garcia2020physics}\\
- & First-Person Hand Action Benchmark\cite{garcia2018first} &\\
MuJoCo & Door opening, in-hand manipulation, & \\
 &  tool use and object
relocation &\\
SUMO\cite{lopez2018microscopic}, METANET\cite{kotsialos2002traffic} &-& \cite{han2022physics} \\
SUMO & - & \cite{wang2022ensuring}\\
CARLA\cite{dosovitskiy2017carla} &-&\cite{udatha2022safe} \\
Gazebo\cite{koenig2004design} & Quadrotor $(IF750A)$&  \cite{huang2023symmetry}\\
%
%
IEEE Distribution & IEEE 33-bus and 141-bus distribution
networks & \cite{chen2022physics}\\
system benchmarks & IEEE 33-node
system & \cite{cao2023physics,chen2022physics} \\
& IEEE 9-bus standard system & \cite{gao2022transient}\\
%
%
- & Custom (COMSOL based) & \cite{alam2021physics} \\
- & Custom (DFT based) & \cite{cho2019physics} \\
- & Custom (based on \cite{vrettos2016experimental})& \cite{gokhale2022physq}\\ 
- & Custom (based on \cite{kesting2007jam}) & \cite{jurj2021increasing}\\
- & Custom &\cite{li2023federated,martin2022reinforcement,dang2022towards}\\
-& Custom&\cite{park2023physics,shi2023physics,yin2023generalizable}\\
- & Custom&\cite{yousif2023physics,yu2023district,zhao2023barrier}\\
- & Custom&\cite{li2019temporal,cohen2023safe,lutter2021differentiable}\\
- & Custom&\cite{wang2023phyllis,chen2023hybrid}\\
Open AI Gym & Custom (based on geometries of Nuclear reactor)  & \cite{radaideh2021physics} \\
MATLAB-Simulink & Custom & \cite{she2023inverter,zhang2022barrier} \\
- & Custom \cite{zimmermann2019puppetmaster}& \cite{mora2021pods}\\
MATLAB & Cruise control & \cite{li2021safe} \\
Pygame & Custom &\cite{zhao2021physics} \\
- & Custom (Unicycle, Car-following) &\cite{emam2022safe} \\
- & Brushbot, Quadrotor (sim)&\cite{ohnishi2019barrier}\\
& Phantom manipulation platform &\cite{lowrey2018reinforcement}\\
Pybullet & 2 finger gripper & \\
& gym-pybullet-drones\cite{panerati2021learning} & \cite{du2023reinforcement}\\
Pybullet & Franka Panda, Flexiv Rizon (also real world robots)& \cite{lv2022sam}\\
NimblePhysics\cite{werling2021fast}, & &\\
Redner\cite{li2018differentiable} (Differentiable sim.) &&\\
&& \\
- & Custom MOCAP  &\cite{chentanez2018physics,peng2018deepmimic,luo2020kinematics}\\
\hline
\hline
\end{tabular}}
\label{table:PIRL_evaluation_Benchmarks}
\end{table}
%
%

\section{Open Challenges and Research Directions}\label{sec:challenges}

\subsection{High Dimensional Spaces}
A large number of real world tasks deals with high dimensional and continuous state and action spaces. One popular method to address this high dimensionality issue is to compress the state space (or action space) vectors into low dimensional vectors. A PI based approach may learn high quality environment representations using deep networks and extract physically relevant low dimensional features from them. 

But learning a compressed and informative latent space from high dimensional continuous state (or action) space still remains a hurdle. Also learning physically relevant representation is still an open problem. Future research should address this issue and try to devise approaches that helps to incorporate or take guidance of underlying physics during representation learning or feature extraction, so as to  make them both informative and physically pertinent. 

\subsection{Safety in Complex and Uncertain Environments}
In the realm of safe reinforcement learning, striking a balance between the complexity of the environment and ensuring safety is always a challenge. Current physics informed approaches uses different control theoretic concepts e.g. CBFs to ensure safe exploration and learning of the RL agent. But these approaches are limited by the approximated model of the system and the prior knowledge about safe state sets. There has been a lot of research for better system identification or model learning through physics incorporation. But most works do not generalize well to different tasks and environments. To summarize, future works should address these crucial research goals: 1) model agnostic safe exploration and control using RL agents in complex and uncertain environments and 
2) devise generalized approach of incorporating physics in data-driven Model learning.

\subsection{Choice of physics prior}
Choice of the physics prior is very crucial for the PIRL algorithm. But such choice is difficult and requires extensive study of the system and may vary extensively from one case to another even in same domains. To enhance efficacy, devising a comprehensive framework with physics information to manage novel physical tasks is preferable rather than dealing with tasks individually. 

\subsection{Evaluation and bench-marking platform}
Currently, PIRL doesn't have comprehensive benchmarking and evaluation environments to test and compare new physics approaches before induction. This limitation makes it challenging to assess the quality and uniqueness of new works. 

Additionally, most PIRL works rely on customized environments related to a particular domain, making it difficult to compare PIRL algorithms fairly. Moreover, PIRL application cases are diverse, and the physics information chosen is specific to a domain, requiring extensive study and domain expertise to understand and compare such works.

\section{Conclusions}
\label{sec:conclusion}
This paper presents a state-of-the-art reinforcement learning paradigm, known as physics-informed reinforcement learning (PIRL). By leveraging both data-driven techniques and knowledge of underlying physical principles, PIRL is capable of improving the effectiveness, sample efficiency and accelerated training of RL algorithms/ approaches, for complex problem-solving and real-world deployment. We have created two taxonomies that categorize conventional PIRL methods based on physics prior/information type and physics prior induction (RL methods), providing a framework for understanding this approach. To help readers comprehend the physics involved in solving RL tasks, we have included various explanatory images from recent papers and summarized their characteristics in Tables~\ref{table:PIRL_Characteristic_MF} and ~\ref{table:PIRL_Characteristics_MB}. Additionally, we have provided a benchmark-summary table \ref{table:PIRL_evaluation_Benchmarks} detailing the training and evaluation benchmarks used for PIRL evaluation. Our objective is to simplify the complex concepts of existing PIRL approaches, making them more accessible for use in various domains. Finally, we discuss the limitations and unanswered questions of current PIRL work, encouraging further research in this area.

\section{Acknowledgment} This research was partly supported by the Advance Queensland Industry Research Fellowship AQIRF024-2021RD4.

\printbibliography 

\end{document}